\documentclass[pdflatex,sn-basic]{sn-jnl}% Basic Springer Nature Reference Style/Chemistry Reference Style

\usepackage{graphicx}%
\usepackage{multirow}%
\usepackage{amsmath,amssymb,amsfonts}%
\usepackage{amsthm}%
\usepackage{mathrsfs}%
\usepackage[title]{appendix}%
\usepackage{textcomp}%
\usepackage{manyfoot}%
\usepackage{booktabs}%
\usepackage{algorithm}%
\usepackage{algorithmicx}%
\usepackage{algpseudocode}%
\usepackage{listings}%
\usepackage{bold-extra}
\usepackage{xurl}
\usepackage{microtype}
\usepackage{tabularx}
\usepackage[table]{xcolor}
\usepackage{xspace}
\usepackage{xltabular}
\usepackage[T1]{fontenc}%
\usepackage{lmodern}

\newcommand{\hatecheck}{\textsc{HateCheck}\xspace}%
\renewcommand{\cite}{\citep}

\begin{document}

\title[Specification overfitting]{Specification Overfitting in Artificial Intelligence}


\author[1,2]{\fnm{Benjamin} \sur{Roth}}\email{benjamin.roth@univie.ac.at}\equalcont{These authors contributed equally to this work.}

\author*[1,3]{\fnm{Pedro Henrique} \sur{Luz de Araujo}}\email{pedro.henrique.luz.de.araujo@univie.ac.at}
\equalcont{These authors contributed equally to this work.}

\author[1,3]{\fnm{Yuxi} \sur{Xia}}\email{yuxi.xia@univie.ac.at}

\author[4]{\fnm{Saskia} \sur{Kaltenbrunner}}\email{saskia.kaltenbrunner@univie.ac.at}

\author[4]{\fnm{Christoph} \sur{Korab}}\email{christoph.korab@univie.ac.at}

\affil[1]{\orgdiv{Faculty of Computer Science}, \orgname{University of Vienna}, \orgaddress{\city{Vienna}, \country{Austria}}}

\affil[2]{\orgdiv{Faculty of Philological and Cultural Studies}, \orgname{University of Vienna}, \orgaddress{\city{Vienna}, \country{Austria}}}

\affil[3]{\orgdiv{UniVie Doctoral School Computer Science}, \orgname{University of Vienna},\orgaddress{\city{Vienna}, \country{Austria}}}

\affil[4]{\orgdiv{Department of Innovation and Digitalisation in
Law}, \orgname{University of Vienna}, \orgaddress{\city{Vienna}, \country{Austria}}}


\abstract{%The data-driven approach to artificial intelligence has been extremely successful when measured by the rate of adoption of artificial neural networks across a wide range of tasks. 
Machine learning (ML) and artificial intelligence (AI) approaches are often criticized for their inherent bias and for their lack of control, accountability, and transparency. Consequently, regulatory bodies struggle with containing this technology's potential negative side effects.
High-level requirements such as fairness and robustness need to be formalized into concrete specification metrics, imperfect proxies that capture isolated aspects of the underlying requirements.
Given possible trade-offs between different metrics and their vulnerability to over-optimization, integrating specification metrics in system development processes is not trivial.
%Recent developments extend the evaluation of ML models with behavioral testing, a concept from software engineering, where failure cases, requirements, and boundary conditions are systematically checked using carefully constructed input-output pairs. In contrast to purely data-driven evaluation, behavioral tests emanate from human prior knowledge and insight into the task and allow for a controlled verification of system properties.
This paper defines \emph{specification overfitting}, a scenario where systems focus excessively on specified metrics to the detriment of high-level requirements and task performance. 
We present an extensive literature survey to categorize how researchers propose, measure, and optimize specification metrics in several AI fields (e.g., natural language processing, computer vision, reinforcement learning). 
Using a keyword-based search on papers from major AI conferences and journals between 2018 and mid-2023, we identify and analyze 74 papers that propose or optimize specification metrics. 
We find that although most papers implicitly address specification overfitting (e.g., by reporting more than one specification metric), they rarely discuss which role specification metrics should play in system development or explicitly define the scope and assumptions behind metric formulations.
%We examine how additional requirements influence the training and evaluation procedures, providing detailed categorical insights and an aggregate overview. The findings include representative and noteworthy discoveries discussed in the study.
}




\maketitle

\section{Introduction}

The classical way \cite{shalev2014understanding} of measuring the performance of predictive systems only on held-out data has been identified as inadequate to fully reflect the complexities of real-world use cases \cite{ribeiro2020beyond}, where it may be required that an algorithm fulfills additional properties that are not sufficiently reflected by reporting average performance metrics such as accuracy on the held-out data.

States, companies, and non-profit organizations have formulated ethical principles and high-level guidelines for AI \cite{fjeld2020principled,hagendorff2020ethics,jobin2019global}. Laws and regulations are being formulated to make adherence to such principles legally binding \cite{wachter2017counterfactual,barocas2016big}. The EU legal framework for AI \cite{veale2021demystifying} requires certification of AI systems on the basis of such laws. However, the details of such regulations are often (implicitly) relegated to standardization organizations (e.g., DIN, ETSI, ISO, NIST\footnote{\url{https://www.din.de/}, \url{https://www.etsi.org/}, \url{https://www.iso.org/}, \url{https://www.nist.gov/}}), and since laws on regulating AI are very recent (or still in the making) there is little experience in how to translate the higher-level principles into low-level evaluation scenarios.

High-level guidelines may be formalized narrowly into concrete specifications and metrics, a process that requires making assumptions---\textit{what} aspects of the underlying goal should be measured and   \textit{how} should they be measured---that can introduce mismatches between high-level principles and their measurements~\cite{jacobs2021measurement}.
%capturing isolated aspects, and it can be expected that only those aspects of the high-level desired outcomes are measured that are formalized. %TODO:CITE?
Moreover, an aspect that has not received enough attention is the question of what role the outcome of an evaluation w.r.t. additional specifications should play in the larger development cycle of AI systems. 
%Once there is more than one metric, many questions do not have a clear answer (optimization can also happen in an indirect, unintended, implicite, way, e.g., by selecting models that fare better wrt specifications).
Most scientific publications (as we will show in our analysis) discussing the use of additional specification metrics for AI systems do not address the question of whether those additional feedback metrics can and should be used during system development and little research has been undertaken to study the effects of considering specification metrics in AI system development.
As a step towards raising awareness of those questions, we provide a comprehensive first overview of papers that consider additional specification metrics, and we catalog the training and evaluation setups that are common when measurements of additional requirements are included in AI and ML scenarios.

There is a long tradition of discussing the potential of unintended consequences and misalignment of goals for AI systems \cite{wiener1960some}. %Wikipedia: In 1960, AI pioneer Norbert Wiener described the AI alignment problem as follows: "If we use, to achieve our purposes, a mechanical agency with whose operation we cannot interfere effectively… we had better be quite sure that the purpose put into the machine is the purpose which we really desire."
Recently, \citet{malik2020hierarchy} discusses the sacrifices and pitfalls of translating open-ended and qualitative questions into quantitative machine learning settings.
However, those works do not discuss how current AI research deals with \emph{competing} objectives and possible feedback loops that include measurements of \emph{additional} specifications.
Similarly, work on testing in ML and AI, e.g., \citet{zhang2022Chasing}, studies different properties that ML and AI systems can be evaluated for and different ways to do it, but there is no guidance on how to integrate \emph{different} metrics in the AI development process.
Work on underspecification in AI has shown how equivalent predictors ---with the same test set task accuracy---can exhibit widely different behaviors on single instances or for properties not reflected in the held-out data. 
\citet{d2022underspecification} remark that one should use specifications to 
%fully specify the desirable outcomes 
enforce desired behaviors and use them to select systems in such cases, but they do not discuss how to resolve disagreements between different specifications and how to prevent overfitting to specifications (see below).

Our study is the first to define \emph{specification overfitting}, 
the case of overfitting to desirable outcomes specified additionally besides the task metric.
Whereas misalignment and underspecification concern the specification of a faulty main objective and the failure to specify additional desirable properties, specification overfitting describes a scenario where specification metrics improve to the detriment of the main task metric or other specifications.
For this, we are the first to comprehensively categorize common practices in scenarios with specification metrics, i.e., different, possibly competing measurements of additional properties besides the main task metric (e.g., accuracy on identically and independently distributed held-out data).
We cover papers from several fields, including NLP, computer vision, and reinforcement learning, and we provide a quantitative and qualitative analysis of the methods and recommendations in those papers.
%We categorize the recommendations 
%For the quantitative analysis, we categorize the approaches in a structured way that allows to make statements of dominating choices on an aggregate level.

We build our analysis on the results of a key-word-based search for papers from DBLP\footnote{\url{https://dblp.org/}} \cite{ley2002dblp} covering the main conferences in NLP, CV, and AI, from 1/2018 -- 7/2023, retrieving those papers that deal with scenarios where additional requirements are measured in addition to a task metric.
We stratified those papers to have equal coverage in 3 groups of application domains (NLP, Vision, Other) by ranking and filtering, keeping the most cited papers.
We keep the resulting set of 74 papers for our in-depth analysis.
We analyze how the additional requirements and specifications are reflected in the training and evaluation procedure described, encoding it in a categorical schema.
We report this fine-grained categorical analysis, together with an aggregate overview, and discuss representative and interesting findings.

%TODO: Quantify the findings
Of all 74 papers that measure an additional specification, 62 papers also attempt to improve on that metric. Forty-eight papers study the effect of the attempt to improve this metric on other metrics (including the task metric).

We find that most papers (59) do not recommend how to use the specifications' feedback in the development process. Of the ones that do, four recommend delegating the decision of how to use specification metrics to an expert, and three recommend using the feedback of specifications for debugging. Only one \cite{pfohl2022Net} provides a concrete recommendation on how a specification should be employed during development to obtain an overall improved system.

Our survey reveals that despite a large body of research on specifications, additional requirements, and their optimization, there is currently no clear recommendation, let alone consensus, on how to use them.
Many works do not even address the concern of over-optimizing specifications.
With increased formalization of regulatory requirements, incentives rise to narrowly follow specifications in artificial intelligence.
Therefore, it is paramount to develop analysis schemes, recommendations, and best practices for developing AI systems with multiple, potentially competing quality metrics and specifications.
%
    
%TODO: Overview of paper
The remainder of this survey is structured in the following way:
Section~\ref{sec:legal} contextualizes how legal frameworks define and regulate specifications.
Section \ref{sec:application_areas} gives an overview of the types of tasks and data included in our study. The common criterion is that additional specifications have been defined for those tasks (in addition to ``success'' on held-out data).
Section \ref{sec:specifications} categorizes the types of additional specifications that have been proposed for measurement.
Section \ref{sec:specification_optimization} outlines approaches for optimizing the different types of specifications.
In Section \ref{sec:survey}, we describe the analysis of papers we reviewed both quantitatively on an aggregate level, and we also discuss a few selected papers to illustrate the variety of work covered in our survey. We summarize our findings in Section \ref{sec:conclusion}.

\section{Specifications and the legal framework}
\label{sec:legal}
% \subsection{Legal framework: The AI Act}
Without proper ethical guidelines and safeguards, AI systems may exacerbate inequalities, further marginalize vulnerable communities, and cause physical or psychological harm.
Responsible AI requires that the design and development of AI systems are aligned with universal values, principles, and international norms \cite{kiden2024Responsible}.
Alignment to ethical values gives rise to a double challenge: the normative challenge of deciding \textit{which} principles should be considered, and the technical challenge of deciding \textit{how} these principles should be encoded in AI systems \cite{iniesta2023human}.
That is, putting AI ethics into practice requires AI regulation and the translation of principles into concrete standards and requirements \cite{bleher2023Reflections}.

Legislators around the world seek to regulate AI technology as the production and deployment of such systems increase rapidly. Such regulation often entails broad terms that leave room for interpretation and need translation into a specification to help developers achieve compliance. The interaction between legal terms and their technical interpretation becomes essential for lawmakers to ensure that systems adhere to the desired principles, and often takes place through the development of recognized standards. While this phenomenon is inherent to technical legislation worldwide, we will highlight the European Artificial Intelligence Act (AI Act)~\cite{EU2024AIAct} as an example because it is the first comprehensive regulation of AI. 

The AI Act is the central European legislation project to establish harmonized requirements for Artificial Intelligence in the Union and attempt a conciliation of the technology with its fundamental values.  
It establishes several requirements that systems must observe to be authorized for the internal market. The AI Act establishes the category of high-risk AI systems in Article 6 and subjects mainly these systems to its requirements. In the final version of the AI Act, General Purpose AI systems are also subject to certain requirements before they can be deployed in the market. It builds---in the spirit of prior European product safety regulation~\cite{EC2008new}---on the established mechanism of European harmonized standards, making visible the interplay between regulatory approaches and technical specifications. 

\paragraph{Enforcing requirements through standardization}
The lawmaker operates--–similar to other technical legislation---with rather broad terms that leave room for interpretation to ensure sufficient flexibility of the acts. In this manner, the AI Act establishes that providers of AI systems have to demonstrate the conformity of their systems to harmonized standards or that the systems have to be tested against ``appropriate standards''. 
As a result, the legislative act alone is not enough for providers to obtain information on the procedure they have to observe. 
Rather – even though only marginally mentioned – the harmonized standards become central sources for establishing conformity. 

Having reliable metrics for measuring certain qualities of AI systems that the legislator (and by extension, society) demands of them becomes essential because the logic of the law transforms these metrics into the determining factor for severe liability questions. 
Liability claims are facilitated by an assumption of causality in the case of non-compliance according to Article 4 (2) of the proposed AI Liability Directive~\cite{EU2022proposal}, and regulatory fines can be administered according to the AI Act.   
On the other hand, for systems compliant with harmonized standards, establishing a liability according to the new product liability regime will become much more difficult, if not impossible, for claimants, because in order to establish the defectiveness of a product, interventions by regulatory authorities and product safety requirements (such as the ones in the AI Act) have to be taken into account. 
Furthermore, Article 10 of the proposed new product liability directive even establishes that liability is excluded if the defectiveness is due to the product's compliance with mandatory regulations (such as the AI Act).

\paragraph{Harmonized standards and the AI Act}
The Act treats favorably the harmonization via the mechanism of Regulation 1025/2012~\cite{EU2012standardization}, in which the Commission issues a standardization request to the standardization organizations. 
A reference to the harmonized standard is then published in the Official Journal of the European Union. Systems complying with such standards will be (rebuttably) assumed to also comply with the related requirements of the AI Act. 
Should this strategy fail, i.e., no harmonized standards be passed, the Act attributes a backup role to the Commission in Article 41. 
In this case, the Commission can become active and adopt ``common specifications''.  

The AI Act treats technical specifications as the central way of determining compliance with the requirements. 
In doing so, it remains vague and delegates the responsibility (and decision-making authority) entirely to executive bodies, namely the standardization organizations. 
%\textbf{TODO Saskia and Christoph: Please check next paragraph}
The Act is governed by two implicit assumptions. It assumes that
\begin{enumerate}
    \item for every requirement imposed on systems, there is a corresponding metric (or several metrics) to measure the fulfillment of the requirement accurately. 
    \item if a system fails to fulfill a specification (e.g., by falling below a pre-determined threshold in a metric), adapting the system to fulfill the specification will not harm its overall performance.
That is because if the system fulfills a specification then it is considered compliant, and the Act does not reference the further impact an adaptation process might have on the system overall. 
\end{enumerate}
In this respect, the Act does not provide procedural guidance on improving a system if it does not satisfy a given metric---nor does it require harmonization bodies to 
provide procedural guidelines.
%adopt such concepts.  

\paragraph{Standardization organizations and decision-making power }
The delegation of decision-making authority by the Commission is recognized by the Court of Justice of the European Union (CJEU) and is not an uncommon tool to support the application of Union legislation.  
When the Commission asks standardization organizations to draft harmonized standards (that will lead to a presumption of conformity under the AI Act), it uses a delegated decision-making authority. 
The standardization organizations work together with the European Union in a public-private partnership. 
A common understanding between standards organizations, the European Commission, and the European Free Trade Association (EFTA) of the principles of this collaboration has been outlined since 1984 and updated in 2003~\cite{EC2003guidelines}.  

In the case of technical metrics for AI systems, this may be problematic because the consequences of the decision (e.g., which standard to adopt) can be more unpredictable than in other technical sectors.
AI systems may narrowly follow the specifications to conform to the standards to the cost of the overall performance of the system, the underlying requirements, or other overlooked relevant aspects. 
% As this paper shows, there are significant divergences in the specifications chosen in current practices. 
Given the considerable potential harms of specification overfitting, it seems worth asking whether this process of delegation to standardization authorities in the context of the AI Act is enough to provide legal protection and means of redress for citizens affected by AI systems and enough legal clarity for developers and providers.

\section{Application areas}
\label{sec:application_areas}
This section summarizes the application areas explored by the papers in our survey.
We group them into categories based on the nature of the data, as different varieties require different specifications.
We give an overview of each category's current state of the art, along with challenges and limitations.

\subsection{Natural language processing}
Natural language processing (NLP) applications use text or speech as input.
They comprise tasks such as sentiment analysis~\cite{socher2013recursive}, machine translation~\cite{bahdanu2015neural}, and named entity recognition~\cite{lample2016neural}.
The state of the art of the area is dominated by transformer-based~\cite{vaswani2017attention} systems trained on massive amounts of text to optimize a language modeling objective~\cite{clark2020ELECTRA,raffel2020exploring,liu2019roberta,devlin2019bert}.
Fine-tuning on instruction datasets~\cite{wei2022finetuned}, optimizing additional objectives~\cite{ouyang2022training}, and scaling up training data and system size allowed the use of language models on complex tasks requiring multi-step reasoning and diverse knowledge~\cite{wei2022emergent}.
Current research investigates the use of language models in areas such as creative writing~\cite{yuan2022wordcraft}, code development~\cite{zan2023large}, education~\cite{Kasneci2023chatGPT}, and medicine \cite{thirunavukarasu2023large}.
Limitations of these systems include the use of non-generalizable heuristics~\cite{tu2020empirical}, generating texts that are biased, hateful, and toxic~\cite{schick2021selfdiagnosis}, and texts that look fluent and plausible but contain falsehoods and misinformation \cite{ji2023survey,lin2022truthfulqa}.

\subsection{Computer vision}
Computer vision (CV) applications process images or videos as input and solve tasks such as image classification~\cite{russakovsky2015imagenet}, segmentation~\cite{minaee2022image}, and face recognition~\cite{he2005face}.
The state of the art is dominated by vision transformers~\cite{kolesnikov2021image} and convolutional neural networks~\cite{lecun1989backpropagation,fukushima1980neocognitron} pretrained on massive image datasets~\cite{he2016deep,szegedy2016rethinking,krizhevskz2012imagenet}.
Vision transformers brought improvements in a wide range of vision tasks~\cite{han2023survey} and gave rise to multimodal systems capable of combining---and producing---visual and text information~\cite{radford2021learning,ramesh2021zero}.
Computer vision has been applied in sensitive areas like healthcare~\cite{esteva2021deep}, surveillance~\cite{sreenu2019intelligent}, and autonomous vehicles~\cite{hu2023planning}.
Despite the good performance on standard benchmarks, there are still technical and ethical limitations such as the lack of robustness to distribution shifts~\cite{bendavid2010theory} and adversarial attacks~\cite{goodfellow2015explaining}, and poor performance on underrepresented demographic groups~\cite{buolamwini2018gender}.

\subsection{Others}
While most of the papers in the survey explored NLP and CV tasks, some investigated tasks that fit other categories.

\textbf{Tabular data}:
Tabular data applications represent input examples as structured records of numerical and categorical features.
Contrary to previous cases, traditional machine learning algorithms such as ensembles of decision trees often still outperform deep learning-based approaches~\cite{borisov2022deep}.
Systems trained on tabular data are applied to a wide range of areas, including sensitive ones such as medical diagnoses~\cite{kononenko2001machine} and financial analyses~\cite{bhatore2020machine}, even though they have been shown to reproduce dataset biases~\cite{angwin2016machine}.

\textbf{Graphs}:
In graph applications, entities and their relationships are represented as nodes and edges in a graph.
Tasks include assigning graphs or nodes to particular classes or predicting links between entities.
State-of-the-art approaches use different variants of graph neural networks (GNNs)~\cite{wu2021comprehensive}, which have been applied to areas such as social network analysis~\cite{fan2019graph} and drug discovery~\cite{xiong2020pushing}.
Examples of current challenges in graph applications are generalization and scalability concerns~\cite{bronstein2017Geometric} and system vulnerability to adversarial attacks~\cite{sun2023adversarial}.

\textbf{Reinforcement learning}:
%In contrast to the previous tasks, which refer to specific input modalities, r
Complex tasks that cannot easily be learned by optimizing local decisions are often modeled in the framework of reinforcement learning.
Here, the problem formulation is that an agent seeks to maximize a reward signal by choosing the optimal action given an environment state~\cite{sutton2018reinforcement}.
The current state-of-the-art methods are based on deep reinforcement learning~\cite{arulkamaran2017deep} and have prominently been applied to video games~\cite{mnih2015human} and robotics~\cite{levine2016end}.
The formulation of the rewards signal is critical, as reinforcement learning systems are vulnerable to reward hacking, where the agent optimizes the reward to the detriment of the task~\cite{skalse2022defining}.
There are also concerns with robustness to noise and adversarial attacks~\cite{lutjens2020certified}.

\section{Specifications}
\label{sec:specifications}
% Specifications are defined in the theory of requirements engineering, where phenomena are grouped into world and machine phenomena~\cite{jackson1995world}.
% For example, if a system is to predict the sentiment polarity of film reviews, world phenomena would include the films, the sentiment of the critics and the interpretation of the annotators.
% Machine phenomena would comprehend the modelling and optimization choices.
% The specifications deal with the intersection of world and machine phenomena: in this example, it could be a dataset of movie reviews annotated with binary sentiment labels.

The requirements engineering framework distinguishes requirements from specifications~\cite{jackson1995world}.
Requirements are concerned with world phenomena, while specifications lie in the intersection of machine and world phenomena.
Requirements can include high-level concepts such as fairness and robustness, which are translated into specifications by defining datasets or metrics intended to assess those properties.

% \subsection{Translating high-level goals into specifications}
% Distinguishing requirements from specifications and programs enables us to isolate problems arising from faults in each of those levels~\cite{kulynych2020POTs}: inadequate descriptions of the problem, inadequate descriptions of the solution and inadequate implementations.
The path between requirements and specifications is perilous: going from the requirement to the specification level requires abstracting away world-only phenomena.
% In the sentiment analysis example, one loses the notion that sentiment is a (multi-dimensional) spectrum not constrained to positive and negative; and that the data annotation may not reflect the actual author's sentiment due to faults in annotators' interpretations (in detecting sarcasm, for example).
% Though this abstraction process is inevitable, one should do so in a way that reflects the system's purposes.
% This is further complicated in the machine learning domain due to data quality
% concerns and the lack of explainability of black-box models~\cite{ahmad2021whatsup}.
% This can be analyzed through a measurement model framework.
The requirements are constructs: unobservable theoretical abstractions that describe phenomena of interest, such as robustness and fairness~\cite{jacobs2021measurement}.
These cannot be measured directly, as they are not observable.
Instead, constructs are \textit{specified} through a measurement model that leverages observable properties, or proxies (e.g., accuracy on a dataset, invariance tests, bias metrics), to infer the construct.
That involves making assumptions about the relevant observable properties and how they relate to the unobservable construct and each other, potentially introducing mismatches between the theoretical understanding of the problem and its operationalization~\cite{jacobs2021measurement}.

% In this paper, we are interested in how optimization can lead to programs that fulfil the specifications but not the requirements~\cite{kulynych2020POTs}.
% We understand this as specification overfitting: we assume a set of specifications and investigate how optimizing on this restricted set of narrow expectations about how an AI system should work can be problematic.
% A better understanding of the process in which problems arise from specific choices of how and what to specify can help fulfil the actual requirements in two ways.
% First, it would inform how to better translate requirements into specifications.
% Second, it could lead to the creation of tools for preventing such over-optimization.

In the rest of this section, we describe aspects of interest to our survey, and how we categorized the surveyed papers w.r.t. different types of sepcifications and other properties.
%specified in our survey's papers and how they are specified.

\subsection{What to specify}
\label{sec:specifications_what}

The papers from our survey measure specifications that we categorize into three groups.

\paragraph{Robustness}
Robustness concerns how well a system works on examples whose distribution differs from the training distribution.

Often, a specific desideratum for robustness is that a small change in the input should lead to no (or only a small) change in the output.
To test these properties, one can either rely on naturally occurring distribution shifts between data sets or create test examples by perturbing the input of examples and requiring stability on the output side~\cite{wang2022measure}.
The first case is a common issue when systems are used in the wild: NLP systems may have to process texts from different genres, dialects, and grammaticality; CV systems may have to process images with different lighting conditions, perspectives, and quality.
The second case are perturbations, changes to the input part of examples designed to systematically test the effect on the predicted output. Perturbations are often used in the context of adversarial attacks~\cite{zhang2020adversarial}, where the aim is to fool the system into changing its prediction with minimal, unperceivable changes to the input.
The more robust a system is, the less such environmental or adversarial changes degrade its performance.

Robustness is addressed directly in the AI Act (Art 15) as one of the central requirements for high-risk AI systems, along with accuracy. 
Article 15 para 1a also explicitly obliges the Commission to encourage the development of industry benchmarks and measurements to determine accuracy and robustness. 
These may differ from the harmonized standards in Article 40, but Article 15 para 1a, with its wording, encourages a system of de-facto industry standards to exist equally besides the harmonized standards. 

Also, before the AI Act, robustness had an extensive tradition as a key aim in developing AI systems. 
In the Trustworthy AI Guidelines~\cite{HLEG2019guidelines}, systems are required to be “lawful, ethical, and robust” to be considered trustworthy. 
This document was already produced in 2019, and followed by a large number of policy initiatives undertaken by the European Commission~\cite{EC2018factsheet}. 
Other actors, such as the OECD, have also picked up the notion of robustness, making it prominent in AI policy also outside and before the AI Act~\cite{OECD2019ai}. Yet, on a policy level, there is no unanimous agreement on its definition, how it can be measured, or the threshold for a system to be considered robust.

\paragraph{Fairness}
Machine learning systems can reflect societal biases in their training data, such as gender and racial stereotypes.

It has been well-documented how deploying such systems has harmed and further marginalized vulnerable communities~\cite{mehrabi2021survey}.
Such harms can be mitigated by enforcing fairness constraints in the system predictions.
There are multiple competing notions of fairness (e.g., individual fairness, equal opportunity, demographic parity, counterfactual fairness), leading to several fairness metrics~\cite{barocas2019fairness} and fairness enhancing methods~\cite{pessach2023review}.

The term ``fairness'' is not used in the AI Act to refer to a distinct quality of systems that the Act requires. 
Nonetheless, the notion of fairness as a key requirement of AI systems is deeply ingrained in policy and commonly used ethical guidelines, which in turn often draw on fundamental rights discourses. 
For instance, the European Commission’s High-Level Expert Group on Trustworthy AI lists four ethical principles for AI systems, derived from fundamental rights. One of these principles is fairness, closely linked to the rights to Non-discrimination, Solidarity and Justice (Art 21 and following in the EU-Charter)~\cite{HLEG2019guidelines}. Fairness as a requirement for AI also appears to have established itself in academia and among practitioners, more than, for instance, the related concepts of equity or justice. 
That can be seen in a wide range of organizations aiming to develop AI fairness checklists (see, for example, \citet{madaio2020coDesigning}).

While the AI Act does not pick up the specific term, Article 10 (2) (fa) requires providers to establish ``appropriate measures to detect, prevent and mitigate possible biases'' in the training, validation, and testing data set. 
Providers must consider metrics and techniques that test for and mitigate biases. 

This obligation to detect and mitigate biases only explicitly applies to the data sets used.
However, it could be construed as to also include mitigation techniques for the system rather than (only) the data sets, so as to further combat possible biased output.

\paragraph{Capabilities}
We define a capability as a fine-grained aspect of desired task behavior.
That includes diverse phenomena such as linguistic~\cite{ribeiro2020beyond}, numerical reasoning~\cite{naik2018stress} and generalization~\cite{lake2018generalization} capabilities.
Capabilities are often evaluated using test suites~\cite{rottger2021hatecheck,ribeiro2020beyond} comprising specific examples that relate to the tested capabilities.

While the AI Act obliges providers of AI systems to list the capabilities of the respective system as a means of transparency, it refers to capabilities as a technical specification primarily in the context of general-purpose AI. Such AI systems are considered to represent systemic risks if they have high capabilities (see Recital 60n). As a first approximation, the Act uses the amount of compute used for training and sets the initial threshold at $10^{25}$ FLOPs. Systems above this threshold are presumed to represent systemic risk. The European legislator furthermore explains in Recital 60n of the AI Act that this threshold of $10^{25}$ FLOPs should be adjusted over time as well as \textit{be supplemented with benchmarks and indicators for system capabilities}, i.e., means of specification other than compute power. The Commission is granted an explicit mandate to amend the threshold for compute and adopt such benchmarks and indicators as specifications that, when met, trigger the presumption of systemic risk. 

Therefore, from a regulatory perspective, capabilities' specifications play a central role in assessing general-purpose AI systems and categorizing those systems according to their possible risks.  

\subsection{How to specify}
\label{sec:specifications_how}
% Let $h_{\theta}: \mathcal{X} \rightarrow \mathcal{Y}$ design a machine learning system parametrised by $\theta$, $\mathcal{D}$ denote a dataset, and $f: h_{\theta} \times \mathcal{D} \rightarrow \mathbb{R}^k$ be an evaluation function that takes the system and the dataset as inputs and returns a set of $k$ values.
We distinguish between two specification categories according to how the measured property is encoded: example-based specifications, with the property encoded by a set of examples, and metric-based specifications, with the property encoded by a dedicated metric.

\paragraph{Example-based specifications}
% Properties of intereset encoded in the examples
  % f(Theta, {X, X_group, …}, {Y, Y_group, …})
  % X’s, Y’s capture properties of interes, f measure agreement to properties of interest
  % Prototypical cases/Checklists
  % Perturbation-based?
  % Rationales
  % Accuracy?
In this scenario, the requirements are validated through input-output examples that correspond in some way to the tested property.
For example, to measure the robustness of a computer vision system, one can compute the accuracy on a set of perturbed samples~\cite{ross2018improving}.
Our survey categorizes example-based specifications into five types:

\textbf{Human-generated}: examples are written (or otherwise composed) by humans. E.g., the examples in the Crowdsourced Stereotype Pairs (CrowS-Pairs)~\cite{nangia2020crows} dataset were created by asking crowdsourced workers to write sentences reflecting (or violating) stereotypes about demographic groups.

\textbf{Pattern-generated}: examples are generated algorithmically through template filling or rules. E.g., the examples in the INequality Theorem (INT)~\cite{wu2020INT} benchmark are generated by a rule-based algorithm that automatically generates theorems.

\textbf{Model-generated}: examples are sampled from or generated by a probabilistic model. E.g., \citet{bartolo2021Improving} used a pre-trained language model to generate synthetic question-answer pairs that improved the robustness of question-answering systems trained on them.

\textbf{Perturbation-based}: examples are generated by perturbing samples from a dataset. E.g., perturbing images from a dataset by adding to each example a vector that changes the system prediction while keeping perturbed and original images indistinguishable by humans~\cite{goodfellow2015explaining}.

\textbf{Selection-based}: examples are selected from existing datasets to focus on the tested property. E.g., splitting graph datasets into a training set with the smallest graphs and a testing set with the biggest graphs to assess size generalization~\cite{buffelli2022SizeShiftReg}.

\paragraph{Metric-based specifications}
In contrast to example-based specifications, metric-based specifications correspond to formalized scores for measuring properties of a prediction algorithm \emph{without} needing additional samples or annotations---the tested property is encoded directly in the metric computation. 
% We refer to such cases as metric-based specifications and to the outputs of $f$ as \emph{aggregate metrics}, since they are typically measures over the entire dataset $\mathcal{D}$.
An example of metric-based robustness specification would be measuring the expected perturbation magnitude needed to fool the system~\cite{jakubovitz2018improving}.

There are many metric-based fairness specifications, which are typically computed by comparing statistics of system predictions conditioned on different demographic groups.
Prototypical examples are equality of opportunity \cite{hardt2016equality}, demographic parity \cite{cotter2019Training}, and group calibration \cite{pfohl2022Net}, each comparing different group statistics: true positive rates, positive prediction rates, and calibration, respectively.
It has been shown that no method can satisfy these fairness conditions simultaneously \cite{kleinberg2017inherent}.
Far from being just a mathematical artifact, the incompatibility of fairness metrics points to the differences in the underlying notions of fairness \cite{barocas2019fairness} and value systems \cite{friedler2021impossibility}.

\subsection{Measuring and improving}
Our survey covers papers that \textit{evaluate} specifications or use methods to \textit{improve} systems regarding specifications.

\paragraph{Evaluation} We say a paper in our survey \emph{evaluates} a specification if it measures it. I.e., the paper either proposes a new method of how to evaluate a specification (e.g., by designing a test suite~\cite{kirk2022Hatemojia} or a metric~\cite{weng2018clever}) or studies a previously proposed specification as part of the evaluation (in the simplest case just reports its outcome).

%the paper studies or proposes a method of how to evaluate a specification (e.g., by designing a test suite~\cite{kirk2022hatemoji} or a metric~\cite{weng2018clever}), or just measures a specification as part of the evaluation.

\paragraph{Improvement} Attempts to improve the specification performance (Sec.~\ref{sec:specOptim}) range from designing a fairness optimization method~\cite{deng2023FIFA} or to employing regularization methods for increasing robustness~\cite{ross2018improving}) in the works included in our survey.

Establishing harmonized standards for optimization methods is not a legal requirement in the AI Act. The Commission may, when issuing a request for standardization to a European standardization organization, ask for standardization of such optimization methods. However, it will not transfer into a legal requirement since the legal text of the AI Act does not provide for an obligation to use certain optimization methods to achieve compliance. From a legal perspective, therefore, achieving harmonized metrics will be a requirement for certain systems; it remains, however, up to system developers how to achieve this. 

\section{Specification optimization}
\label{sec:specOptim}
\label{sec:specification_optimization}

Specifications indicate the (mis-)alignment with or degree of fulfillment of specific properties or proxies of capabilities.
While some works on specifications for AI system \cite{ribeiro2020beyond,nangia2020crows} claim that the additional metrics measuring the fulfillment of specifications should only be used as an insight into existing system behavior, it is also possible to use this feedback for optimizing the measured system properties \cite{liu2019Inoculation,bartolo2021Improving}.

Therefore, the goal in research on specifications ranges from the view that no optimization of specification metrics should be attempted to the view that these metrics can be used for system development.
Under the latter view, specification metrics can be helpful to compare and select different settings w.r.t. performance on this specification, use the specification as part of a loss function during optimization, or even to specifically design algorithms for improving performance on a specification.

\subsection{Specification optimization strategies}
\label{sec:specification_optimization_strategies}

We categorize strategies for specification optimization into the following groups of approaches, depending on how direct the influence of the specifications is on the resulting system:

\textbf{No optimization.} Different settings (system types, hyper-parameter choices), each of which is not specifically directed to improve the desired property, are compared w.r.t. performance on the specification metric. This can give guidance as to which of the settings performs better w.r.t. the measured specification (and which system could be chosen if specification performance was prioritized)---but neither the systems themselves nor the training process are targeted at optimizing the metric. For example, the \emph{CheckList} approach \cite{ribeiro2020beyond} provides a detailed analysis of failure categories for sentiment analysis, duplicate question detection, and machine comprehension but does not suggest using the outcome of this analysis for system improvement.

\textbf{Direct optimization.} The specification metric that measures the property of interest is a direct target in optimizing the AI system. This could be a term in the loss function corresponding to the measured quantity or the inclusion of training examples that, by construction, directly reflect the evaluation logic or come from the same pool of examples used to measure the desired property. 
For example, the \emph{FIFA} approach \cite{deng2023FIFA} uses a combined fairness and accuracy loss during optimization. \emph{Inoculation by fine-tuning} uses examples from \emph{challenge sets}, specifically constructed data sets for testing phenomena in natural language inference and question answering, as additional training data.

Direct adjustments of the behavior of an AI system, such as extending prompts with in-context examples that correspond to the evaluation setting \cite{levy2023Diverse}, also fall into this category. If improvement strategies are directly inspired by a specific way a property is measured (rather than the property in an abstract sense or an alternative way of specifying the property), they also count as a direct attempt to improve, even if assumptions and approximations are made.

% FIFA: Making Fairness More Generalizable in Classifiers Trained on Imbalanced Data.
%-> combined fairness and accuracy loss

\textbf{Indirect optimization.} As before, an additional property (apart from performance on the main task) is a target in optimization. However, the optimized property is not the specification but a property that is assumed to be related. For example, regularization strategies could be employed for improving the robustness of the system, even if the exact regularization term does not directly follow from the mathematical formulation of the robustness metric \cite{jakubovitz2018improving,ross2018improving}. In other words, the improvement strategy relates to the desired underlying property but not directly to the metric used to measure it.

\subsection{Specification optimization evaluation}
\label{sec:specification_optimization_analysis}
Specification optimization may impact system performance not only considering the optimized property but also the main task performance and other specifications.
An ideal specification optimization strategy would improve both task performance (as measured by an assumed to be i.i.d.\ test set) and better align the system to the high-level principle encoded by the specification.
However, there are possible unintended consequences of specification optimization.
Mismatches between specifications, their underlying goals, and task performance may lead to the deterioration of system performance in unforeseen ways.

\paragraph{Evaluation metrics}
When evaluating specification optimization, it is critical that the evaluation scheme is constructed to reveal such system degradation---depending on which metrics are considered for evaluation, some of the failure cases may be obfuscated. Metrics either measure performance on the main task or on additional specifications.

% This section delves into the critical aspect of evaluating model performance and robustness while mitigating the risks associated with specification overfitting. 
% We highlight specification overfitting as a major concern, as it can lead to an inaccurate assessment of a model's capabilities and, subsequently, its real-world utility. 
% We consider the following two key points to characterize awareness of specification optimization in the surveyed papers.

% \subsubsection{Reporting Task Metric}

% Reporting the main task metrics is the basis for monitoring a potential negative impact of specification optimization. Therefore,  we distinguish between cases whether the standard metrics for the main task are reported in a paper, even if it's relegated to a footnote or an appendix.
%TODO:BEN:The following subsection is confusing, re-write

\textbf{Task metric.} The task metric is the main measure of system performance, often a correctness metric (e.g., accuracy, f-score) computed on a held-out test set.
% These metrics can provide a common ground for researchers to evaluate and compare different models or algorithms. 
Reporting the task metric can reveal whether a specification optimization strategy degraded general system performance.
% However, if one proposed method mainly focuses on improving specification properties, reporting standard metrics can monitor the potential negative impact of the specification optimization. 
For example, papers that propose methods to improve robustness to adversarial attacks may report the accuracy for the unperturbed test set to verify that the method preserves system performance on clean samples \cite{rebuffi2021Data, hendrycks2019Usinga, xie2019Feature}. 

\textbf{Specification metric.}
The evaluation scheme may include a range of specification metrics.
This can happen by reporting alternative formulations of specifications with the same underlying goal (e.g., measuring several fairness metrics \cite{cotter2019Training}) or specifications that capture different requirements (e.g., assessing robustness to distribution shifts and system calibration \cite{hendrycks*2019AugMix})
% Reporting only standard metrics may not fully capture the nuances or specific requirements of the particular task.
% Property metrics offer valuable insights and contribute to a more comprehensive evaluation of a proposed method. Some papers focus on evaluating the desired properties of different methods, or models can only report the property metric.
% This is because they use the specific property as the only variable to achieve a comparable evaluation. E.g.,  MILP \cite{tjeng2019Evaluating} evaluates the robustness of different neural networks by the adversarial accuracy to different extend of image perturbation.

\paragraph{Specification overfitting analysis}
% In addition to considering the reporting of task metrics, we also note to what extent the surveyed papers attempted an analysis of specification overfitting.
The term \textit{overfitting} describes the case in which an AI system learns features that arise from noise and data variance rather than learning the underlying data distribution~\cite{Webb2010}.
Traditionally, a model is said to have overfitted when it has low train error but high test error \cite{acena2022Minimally}, though \textit{overfitting} is also used to denote other 
types of over-optimization that can lead to
unwanted drops in performance, such as those due to distribution shift and test set reuse \cite{roelofs2019meta}.

Specification overfitting occurs when a specification optimization strategy improves system performance w.r.t the optimized metric but degrades system performance w.r.t. the task metric or other specification metrics.
We categorize evaluation schemes based on the metrics they include and their ability to detect specification overfitting.

% To draw a more detailed picture, we categorize the extent of overfitting analysis based on the result reported for specification metrics and task metrics. 

%In our analysis, we categorize the extent of overfitting based on the result reported for specification metrics and task metrics. The categorization includes the following:

\textbf{No overfitting analysis.} If the evaluation scheme includes only one specification and/or task metric, we consider that there is no specification overfitting analysis.
Reporting only one specification metric does not account for possible effects on other specification metrics---it has been shown that optimizing a set of specification metrics can have catastrophic consequences on other specifications \cite{luzdearaujo2023crossFunctional}.
While reporting the task metric accounts for the overall impact on task performance, it may obfuscate unintended consequences.
For example, the task metric may not significantly change if system behavior improves a little for common cases but degrades a lot for rare ones \cite{liu2021Just}.

% reporting only one metric can potentially lead to overfitting issues, as it may encourage the development of models that are overly tuned to that specific metric. E.g., Inoculation \cite{liu2019Inoculation} evaluates its capabilities (e.g., numerical reasoning, logical reasoning) with only one dataset (task metric) and one specification metric from each property and thus provides no specification overfitting analysis. In addition, some papers that do not attempt to improve the desired specification performance can also lack overfitting analysis. This means these papers only report the static performance of one model and cannot track the potential overfitting issue of the model. RobustBench \cite{croce2021RobustBench}, which mainly aims to provide a standardized benchmark for evaluating the robustness of different models, offers no overfitting analysis, as it does not propose a new method for improving model robustness.   

\textbf{Cross-specification analysis.} 
This comprises evaluation schemes that report at least two specification metrics. Examples of this include reporting the performance for alternative formulations of a specification (e.g., different attack types for adversarial robustness \cite{li2023squeeze,dapello2022Aligning,cheng2022cat}) or evaluating specifications for different requirements (e.g., capability of handling negations and robustness to word overlap in natural language inference \cite{naik2018stress}).
The former guards against narrowly adapting to the specification to the detriment of the underlying requirement.
The latter accounts for possible negative interactions between different requirements.
% (e.g., specification and task metric, or two specification metrics). ????
    
\textbf{Task performance analysis.} 
This describes evaluation schemes that go beyond reporting a single task metric and examine the effect on task performance more deeply. This can involve comparing performance on relevant subgroups of the task data (e.g., reporting the worst group accuracy in addition to the dataset average performance \cite{zhang2022CorrectNContrast, liu2021Just}), or evaluating task performance on additional (assumed to be i.i.d.) test sets from the same task (e.g., \citet{chen2022Pareto}).
These measures can provide a more reliable assessment of the impact of specification optimization on task performance.

\textbf{Comprehensive overfitting analysis.} 
This category covers evaluation schemes that combine cross-specification and task performance analysis (e.g., \cite{pfohl2022Net}). 
By considering multiple specification metrics and deeply examining task performance, such evaluation schemes may identify failure cases of specification optimization and prevent specification overfitting.

\section{A survey of specification overfitting}
\label{sec:survey}
This section presents our survey of specification overfitting.
We sample and analyze papers that propose methods to improve or measure specifications.
The goal is to create an overview of how the research community has dealt with the specification overfitting issue in recent years.

\subsection{Method}
\textbf{Paper collection.}
By keyword search, we collect papers from the DBLP\footnote{\url{https://dblp.org/}} database.
We restrict our search to major conferences and journals on natural language processing, computer vision, and machine learning.\footnote{AAAI, ACL, COLING, Computational Linguistics, CoNLL, CVPR, EACL, ECCV, EMNLP, FAccT, ICCV, ICLR, ICML, IJCAI, NAACL, NeurIPS, and TACL.}
We used the following keywords:
\begin{itemize}
  \item test suite
  \item behavioral$\mid$behavioural$\mid$functional$\mid$stress + test (4 searches).
  \item challenge + set$\mid$dataset (2 searches).
  \item diagnos$\mid$evaluat$\mid$benchmark$\mid$test$\mid$assess$\mid$improv$\mid$increas$\mid$train$\mid$optimi + \{property\} (45 searches).
\end{itemize} 
Where \{property\} corresponds to fair$\mid$robust$\mid$generalis$\mid$generaliz$\mid$capabilit and refers to specifications for fairness, robustness, generalization, and specific capabilities.
Our first collection round happened on December 12, 2022, returning 950 papers.
We did a second round on August 25, 2023, to improve recall for papers from 2022 and add papers from 2023.
That returned 222 more papers.

\textbf{Filtering.} 
First, we restrict the papers to those published in 2018 at the earliest, yielding 1172 papers.
We then examined all abstracts to assess if they fit our inclusion criteria---papers that propose a method to improve or evaluate a specification.
We judged 442 papers as relevant.
We assigned each of them to at least one application area: NLP, CV, or others.

\begin{figure}
  \centering
  \includegraphics[width=.45\linewidth]{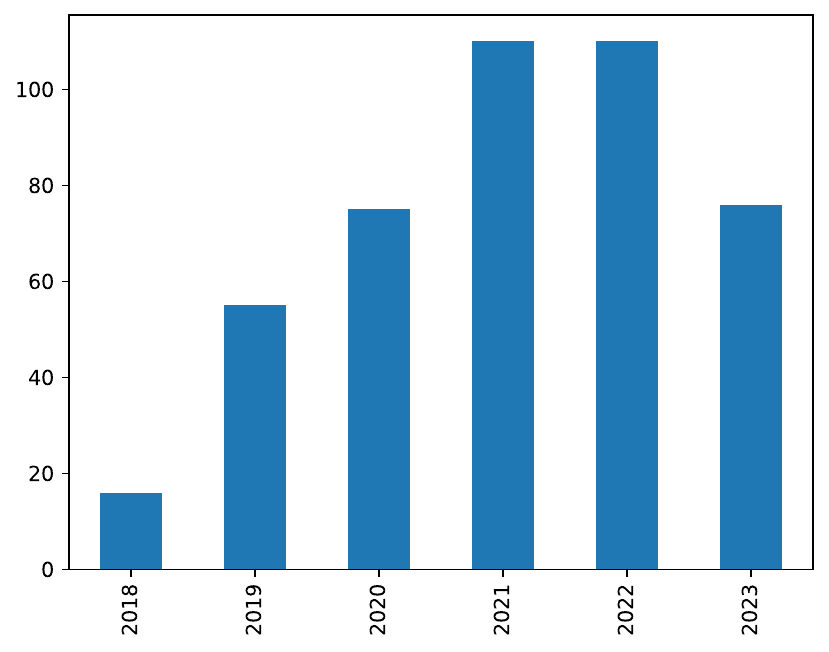}
  \includegraphics[width=.45\linewidth]{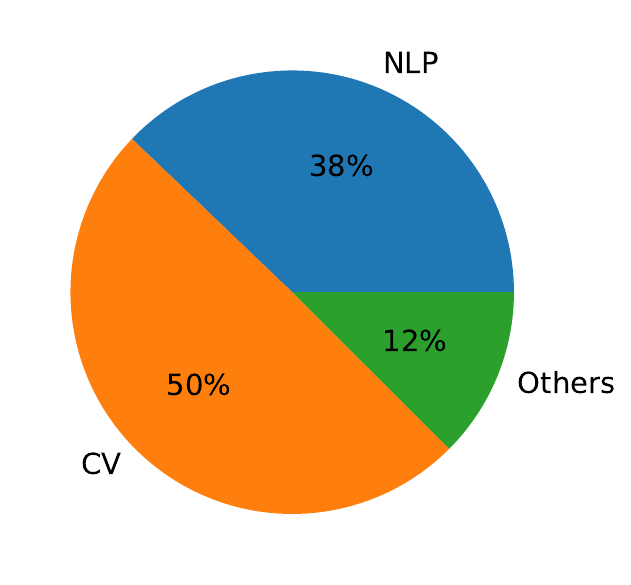}
  \caption{Number of relevant papers by year (left) and application area (right).}
  \label{fig:paperCounts}
\end{figure}

Fig. \ref{fig:paperCounts} shows the number of papers by year and application area.
Interest in measuring and improving specifications seems to be on an upward trend, considering we only sampled papers up to July 2023.
Computer vision is present in about half of papers, followed by natural language processing. 
Only 12\% of the papers explore other application areas (e.g., graph and tabular data).

As a last step, we select the five most cited papers\footnote{As reported in \url{scholar.google.com}.} from each application area for each year.
That was done to keep the workload of analyzing papers manageable while keeping impactful papers and maintaining coverage of different years and application areas.
Due to some papers covering more than one application area and years with fewer than five samples for a given application area, we ended up with 79 papers for annotation.

\textbf{Analysis.}
We read the filtered papers to collect information for the fields in Table~\ref{tab:annotation}.
In this step, we found that five papers did not meet the inclusion criteria, resulting in a final pool of 74 papers.
Table~\ref{tab:annotation} maps the analysis criteria discussed in the different sections of our article to the categories used in the structured analysis, and Table~\ref{tab:allPapers} presents the results of the structured analysis.

\begin{table}[tb]
    \footnotesize
    \centering
    \caption{Paper analysis fields and descriptions.}
    \label{tab:annotation}
    \rowcolors{2}{gray!20}{white}
    \begin{tabularx}{\textwidth}{lX}
        \toprule
        \rowcolor{white}
        Field &Description \\\midrule
        
        Application \textbf{Area}, see Section \ref{sec:application_areas} & The field which the AI/ML application falls in: Natural language processing (\textbf{NLP}), computer vision (\textbf{CV}), tabular data (\textbf{TAB}), graphs (\textbf{GRAPH}), r reinforcement learning (\textbf{RL}).\\

        Specification (\textbf{Spec.}), Sec. \ref{sec:specifications_what} & The specification the paper intends to measure or improve: robustness (\textbf{R}), fairness (\textbf{F}), or capabilities (\textbf{C}). \\
        
        Evaluation (\textbf{Eval.}), Sec. \ref{sec:specifications} &Whether the paper measures a specification (\textbf{\checkmark}). \\

        Example or metric-based (\textbf{Ex/M}), Sec. \ref{sec:specifications_how} & Whether the specification is measured using additional examples (\textbf{e}) or on the same examples as the main task but using a different metric (\textbf{m}). \\
        
        \textbf{Type} of example, Sec. \ref{sec:specifications_how} & If the evaluation is example-based, how the examples are created. We categorize examples into: handcrafted by humans (\textbf{h}), pattern-generated (\textbf{pat}), sampled from a probabilistic model (\textbf{prob}), obtained by perturbing dataset examples (\textbf{per}), or obtained by selecting dataset examples (\textbf{s}). \\

        Improvement (\textbf{Imp.}), Sec. \ref{sec:specification_optimization} &Whether the paper experiments with improving a specification metric (\textbf{\checkmark}). \\
        
        Improvement strategy (\textbf{Imp. Str.}), Sec. \ref{sec:specification_optimization_strategies} &Whether the improvement strategy is based on directly (\textbf{d}) optimizing the specification metric (or a proxy) or indirectly (\textbf{i}) through other means (e.g., regularization). \\

        Reports task (i.i.d.) metric (\textbf{Task M.}), Sec. \ref{sec:specification_optimization_analysis}
 &Whether the paper reports a correctness metric for a standard dataset  (\textbf{\checkmark}). \\

        Overfitting analysis (\textbf{Ov. An.}), Sec. \ref{sec:specification_optimization_analysis} & Whether the paper reports other additional (i.e., more than one) specification metrics (\textbf{o}) and/or studies the effect on task performance (\textbf{t}) in detail. \\
        
        Scope/limitations (\textbf{S/L}), Sec. \ref{sec:limitation} &Whether the paper explicitly discusses the method's scope, e.g., intended use, limitations, assumptions (\textbf{\checkmark}). \\
        
        Recommendation category (\textbf{Rec.}), Sec. \ref{sec:analysis_of_recommendations} & If the paper offers a recommendation on how to integrate the specification metric or the improvement method to the system development process, we categorize it into vague (\textbf{V}), delegating (\textbf{Del}), (not) additional data (\textbf{($\neg$)D}), debugging (\textbf{Deb}) and concrete (\textbf{C}). \\
        \rowcolor{white}
        \bottomrule
        \end{tabularx}
  \end{table}

\begin{table}[!tbph]
\tiny
  \centering
  \setlength{\tabcolsep}{2pt}
  \rowcolors{2}{gray!20}{white}
  \caption{Structured analysis of survey papers.}
  \label{tab:allPapers}
  \begin{tabularx}{\textwidth}{lllXXlXXXXXl}
        \toprule
        \rowcolor{white}
        Paper & Area & Spec. & Eval. & Ex/M & Type & Imp. & Imp. str. & Task M. & Ov. An. & S/L & Rec. \\
        \midrule
        \cite{bartolo2021Improving} & NLP & R & \checkmark & e & prob & \checkmark & i & \checkmark & o &  & D \\
\cite{black2020FlipTest} & TAB & F & \checkmark & e & prob &  &  &  &  & \checkmark & V+Del \\
\cite{buffelli2022SizeShiftReg} & GRAPH & C & \checkmark & e & s & \checkmark &  &  &  & \checkmark &  \\
\cite{chen2019Fairness} & TAB & F & \checkmark & m &  &  &  &  &  & \checkmark &  \\
\cite{chen2022Pareto} & NLP/CV & C & \checkmark & e & s & \checkmark & i & \checkmark & t &  &  \\
\cite{cheng2019Evaluating} & NLP & R & \checkmark & e & prob & \checkmark & d & \checkmark & o &  &  \\
\cite{cheng2020Seq2Sick} & NLP & R & \checkmark & e & prob &  &  &  &  &  &  \\
\cite{cheng2022cat} & CV & R & \checkmark & e & per & \checkmark & i & \checkmark & o &  &  \\
\cite{clarysse2022Why} & CV & R & \checkmark & e & per & \checkmark & i+d &  & o &  &  \\
\cite{coston2020Counterfactual} & TAB & F & \checkmark & m &  &  &  & \checkmark &  &  &  \\
\cite{cotter2019Training} & TAB & F & \checkmark & m &  & \checkmark & d & \checkmark & o &  &  \\
\cite{croce2021RobustBench} & CV & R & \checkmark & e & per &  & d & \checkmark &  & \checkmark & V \\
\cite{dapello2022Aligning} & CV & R & \checkmark & e & per & \checkmark & i & \checkmark & o & \checkmark &  \\
\cite{deng2023FIFA} & TAB & F & \checkmark & m &  & \checkmark & d+i & \checkmark &  & \checkmark &  \\
\cite{elkahkyChallengeSetMethods2018} & NLP & C & \checkmark & e & s & \checkmark & d+i & \checkmark & o &  &  \\
\cite{fatemi2023Improving} & NLP & C & \checkmark & e & h+s & \checkmark & i & \checkmark &  & \checkmark &  \\
\cite{gan2019Improving} & NLP & R & \checkmark & e & h+prob & \checkmark & d & \checkmark &  &  &  \\
\cite{geirhos2018ImageNettrained} & CV & R & \checkmark & e & per & \checkmark & i & \checkmark & o &  &  \\
\cite{gowal2021Improving} & CV & R & \checkmark & e & per & \checkmark & i & \checkmark & o & \checkmark &  \\
\cite{guo2022Comprehensive} & RL & R & \checkmark & e & per &  &  & \checkmark &  &  &  \\
\cite{guoSparseDNNsImproved2018} & CV & R & \checkmark & m &  & \checkmark & i & \checkmark &  &  &  \\
\cite{havasi2020Training} & CV & R & \checkmark & e & per & \checkmark & i & \checkmark & o &  &  \\
\cite{hendrycks*2019AugMix} & CV & C+R & \checkmark & e & per & \checkmark & i & \checkmark & o &  &  \\
\cite{hendrycks2018Benchmarking} & CV & R & \checkmark & e & per & \checkmark & d+i & \checkmark & o &  &  \\
\cite{hendrycks2019Usinga} & CV & R & \checkmark & e & per & \checkmark & i & \checkmark & o &  &  \\
\cite{jakubovitz2018improving} & CV & R & \checkmark & m &  & \checkmark & i & \checkmark &  &  &  \\
\cite{jung2022Reweighting} & TAB & F & \checkmark & m &  & \checkmark & d & \checkmark &  &  &  \\
\cite{karpukhin2019Training} & NLP & R & \checkmark & e & per & \checkmark & d & \checkmark & o & \checkmark &  \\
\cite{kirichenko2022Last} & NLP/CV & R & \checkmark & e & s+per & \checkmark & i & \checkmark & o &  &  \\
\cite{kirk2022Hatemojia} & NLP & C & \checkmark & e & pat & \checkmark & d & \checkmark & o & \checkmark & D \\
\cite{komiyamaNonconvexOptimizationRegression2018} & TAB & F & \checkmark & m &  & \checkmark & d & \checkmark &  &  &  \\
\cite{lee2022Fast} & TAB & F & \checkmark & m &  & \checkmark & d & \checkmark & o &  &  \\
\cite{levy2023Diverse} & NLP & C & \checkmark & e & s & \checkmark & d+i & \checkmark & o+t & \checkmark &  \\
\cite{li2023squeeze} & CV & R & \checkmark & e & per & \checkmark & i & \checkmark & o &  &  \\
\cite{li2023Accurate} & TAB & F & \checkmark & m &  & \checkmark & d & \checkmark & o &  &  \\
\cite{liang2022Efficient} & RL & R & \checkmark & e & per & \checkmark & d & \checkmark & o & \checkmark &  \\
\cite{liu2019Inoculation} & NLP & C & \checkmark & e & pat+h & \checkmark & d & \checkmark &  & \checkmark &  \\
\cite{liu2021Just} & NLP/CV & R & \checkmark & e & s & \checkmark & i & \checkmark & t &  &  \\
\cite{ma2022Are} & NLP/CV & R & \checkmark & e & s & \checkmark & i & \checkmark & o & \checkmark &  \\
\cite{madrasPredictResponsiblyImproving2018} & TAB & F & \checkmark & m &  & \checkmark & d & \checkmark & t &  &  \\
\cite{min2020Syntactic} & NLP & R & \checkmark & e & pat+per & \checkmark & d & \checkmark & o &  &  \\
\cite{mishler2021Fairness} & TAB & F & \checkmark & m &  & \checkmark & d & \checkmark &  & \checkmark &  \\
\cite{naik2018stress} & NLP & C & \checkmark & e & pat+per & \checkmark & d & \checkmark & o &  & V+Del \\
\cite{nangia2020crows} & NLP & C & \checkmark & e & h &  &  &  &  & \checkmark & Deb+$\neg$D \\
\cite{narasimhan2019Optimizing} & NLP/TAB & F & \checkmark & m &  & \checkmark & d & \checkmark &  &  &  \\
\cite{petersen2023assessing} & TAB & F & \checkmark & m &  &  &  &  &  & \checkmark & V+Del \\
\cite{pfohl2022Net} & TAB & F & \checkmark & m &  & \checkmark & d & \checkmark & o+t & \checkmark & C  \\
\cite{qiu2022Improving} & NLP & C & \checkmark & e & s & \checkmark & d & \checkmark & o & \checkmark & D \\
\cite{rahmattalabi2020Fair} & TAB & F & \checkmark & m &  & \checkmark & d & \checkmark &  & \checkmark &  \\
\cite{rebuffi2021Data} & CV & R & \checkmark & e & per & \checkmark & i & \checkmark & o &  &  \\
\cite{ribeiro2020beyond} & NLP & C+F+R & \checkmark & e & pat+per &  &  & \checkmark & o & \checkmark & Deb \\
\cite{roh2021Sample} & TAB & F+R & \checkmark & m &  & \checkmark & d & \checkmark &  & \checkmark &  \\
\cite{ross2018improving} & CV & R & \checkmark & e & per & \checkmark & i & \checkmark & o &  &  \\
\cite{rottger2021hatecheck} & NLP & C & \checkmark & e & h+pat &  &  & \checkmark & o & \checkmark & Deb \\
\cite{ruis2020Benchmark} & NLP & C & \checkmark & e & pat & \checkmark & d &  & o &  &  \\
\cite{schneider2020Improving} & CV & R & \checkmark & e & per & \checkmark & d+i & \checkmark & o & \checkmark & V \\
\cite{sehwag2022Robust} & CV & R & \checkmark & e & per & \checkmark & i &  & o &  &  \\
\cite{sinhaCertifyingDistributionalRobustness2019} & CV & R & \checkmark & m &  & \checkmark & d & \checkmark & o &  &  \\
\cite{sun2020TestTime} & CV & R & \checkmark & e & per+s & \checkmark & i & \checkmark & o &  & V \\
\cite{taskesen2021Statistical} & TAB & F & \checkmark & m &  & \checkmark & i & \checkmark &  & \checkmark & V+Del \\
\cite{tjeng2019Evaluating} & CV & R & \checkmark & e & per &  &  &  &  &  &  \\
\cite{wang2019Improving} & CV & R & \checkmark & e & per & \checkmark & i & \checkmark & o &  &  \\
\cite{wang2020InfoBERT} & NLP & R & \checkmark & e & h+pat+per & \checkmark & d+i & \checkmark & o &  &  \\
\cite{wang2020Robust} & TAB & F & \checkmark & m &  & \checkmark & d & \checkmark &  &  &  \\
\cite{wang2022Identifying} & NLP & R & \checkmark & e & s & \checkmark & d &  & o & \checkmark &  \\
\cite{wangRobustMachineComprehension2018} & NLP & R & \checkmark & e & per & \checkmark & d+i & \checkmark & o &  &  \\
\cite{weng2018clever} & CV & R & \checkmark & m &  &  &  &  &  &  &  \\
\cite{wu2020INT} & GRAPH & C & \checkmark & e & pat & \checkmark & i &  & o & \checkmark & V+Del \\
\cite{xie2019Feature} & CV & R & \checkmark & e & per & \checkmark & i & \checkmark & o & \checkmark &  \\
\cite{zhang2019Stable} & CV & R & \checkmark & e & per & \checkmark & i & \checkmark & o &  &  \\
\cite{zhang2022Chasing} & GRAPH & R & \checkmark & e & per & \checkmark & d+i & \checkmark &  &  &  \\
\cite{zhang2022CorrectNContrast} & NLP/CV & R & \checkmark & e & s & \checkmark & i & \checkmark & t &  &  \\
\cite{zhang2022MEMO} & CV & R & \checkmark & e & per & \checkmark & i & \checkmark & o & \checkmark &  \\
\cite{zhuo2023Robustness} & NLP & R & \checkmark & e & per & \checkmark & d & \checkmark & o & \checkmark &  \\
        \bottomrule
        \rowcolor{white}
    \end{tabularx}
\end{table}

\subsection{Quantitative results}

\textbf{Evaluation and improvements.}
Due to the inclusion criteria, all of the papers evaluate a specification.
Sixty-two papers explore specification optimization strategies.
Of these, 26 include only direct methods, 27 only indirect methods, and nine combine direct and indirect means of improvement.

\textbf{Specification.}
Robustness was the most common specification, with 44 papers, followed by fairness (19) and other specific capabilities (15).
Each specification was conceptualized in many distinct ways.
Robustness was understood as measures of system performance under adversarial attacks~\cite{guoSparseDNNsImproved2018}, distribution shifts~\cite{hendrycks2018Benchmarking}, inference heuristics~\cite{min2020Syntactic}, different subpopulations~\cite{liu2021Just}, missing modalities~\cite{ma2022Are}, and question paraphrasing~\cite{gan2019Improving}.
Fairness measures were very diverse, including, for example, equalized odds~\cite{wang2020Robust}, demographic parity~\cite{coston2020Counterfactual}, equal opportunity~\cite{cotter2019Training}, individual fairness~\cite{black2020FlipTest}, and calibration by group~\cite{petersen2023assessing}. 
Capabilities included generalization~\cite{wu2020INT}, calibration~\cite{hendrycks*2019AugMix}, handling of linguistic phenomena~\cite{naik2018stress}, level of bias~\cite{nangia2020crows}, reasoning~\cite{liu2019Inoculation}, and task-specific capabilities, e.g., recognizing emoji-based hate~\cite{kirk2022Hatemojia}.

\textbf{Example vs. metric-based specifications.}
Example-based specifications were the most common, with 53 papers.
The majority were perturbation-based (32), followed by selection of specific dataset examples (12), pattern-generated (9), human-generated (6), and model-generated (5). 

\textbf{Overfitting analysis.}
Most papers (61) report main task correctness and 26 papers do not include a specification overfitting analysis.
Of the papers that include specification overfitting results, 42 report cross-specification results, four present task performance analyses, and only two are categorized as presenting a comprehensive overfitting analysis.
If we consider only the papers with improvement methods (62), six do not report main task correctness, and 16 do not present any specification overfitting analysis.

\textbf{Scope and recommendations.}
Only 30 of the papers explicitly discuss the scope or limitations of the proposed specification or optimization strategy (we do not consider the mentioning of limitations w.r.t. other aspects).
Further, only 15 papers discuss how the specification should or should not be used during system development or contextualize the role of the specification given the main task and other specifications.

\textbf{Discussion.}
The analyzed papers often implicitly guard against some specification overfitting pitfalls by reporting either the main task performance or metrics for other specifications.
Evaluating main task performance singles out methods that improve a specification to the detriment of general correctness (e.g., always predicting the same outcome is robust against adversarial attacks but has poor task performance).
Measuring performance on other specifications checks whether a method has improved a particular specification to the detriment of others (e.g., improving how a hate speech detector improves for a given demographic while decreasing performance for another).

However, most papers do not explicitly discuss the scope of the proposed method and even fewer contextualize its role in the system development process.
Describing the scope is a way to prevent more insidious pitfalls, such as taking good specification performance as a guarantee of system quality (e.g., claiming that a system with a good performance on a specific fairness metric is \emph{fair}) or not using the method as it was originally intended (e.g., fine-tuning a language model on a dataset intended for evaluation of bias only).

\subsection{Scope and limitations analysis}
\label{sec:limitation}
We identified whether papers explicitly stated the scope and limitations of the proposed specification or improvement method.
% In the pursuit of advancing scientific knowledge, it is imperative to scrutinize not only the strengths but also the limitations of the proposed methodologies. 
%\paragraph{Explication of limitation}
We consider that a paper explicitly discusses the scope and limitations of the proposed specification measure or optimization strategy if
%the relationship to the proposed approach is mentioned, and 
it describes the cases for which the method applies or for which ones it does not. 
This can be done, for example, by discussing the assumptions underlying the proposed method, by contrasting it with alternative formulations, or by discussing in which context the method should be used.
If the paper does not include such discussions, we consider that the scope and limitations were not made explicit.

% For methodologies intertwined with mathematical proofs, the assumptions governing these proofs become pivotal. It is not enough to merely present proofs and formal assumptions, but a rigorous discussion on the applicability and context of these assumptions is necessary for making the limitations of theoretical results explicit.
Scope and limitations are important not only from a practical and scientific position but also from a legal one.
Suppose a harmonized standard does not fully cover a legal requirement.
In this case, compliance with the standard will not establish the (full) presumption of conformity.
Considering this, systems might have to comply with several harmonized standards to obtain a presumption of conformity with one of the high-level legal requirements, such as, for example, robustness. 

\paragraph{Examples of scope and limitations and counterexamples}
In the following, we show types of explicitly defined scope as described in the included papers. 
We also present counterexamples that illustrate how the scope or limitations of a specification or improvement method are not sufficiently defined.
% that falls outside the defined scope. 
%This counterexample serves as a cautionary illustration, highlighting the instances where the proposed method by the paper may not apply. 
%We sampled four types of explicit representative limitations and lack thereof stated in the papers below:

    \textbf{Scope of application context.} \citet{ribeiro2016should} state that their proposed test suites can only account for behavioral (input-output) issues but not non-behavioral issues such as noisy and biased training data, lack of interpretability or security issues. In contrast, some papers do not explicitly restrict the context for applying the proposed method. 
    For example, papers examining adversarial robustness \cite{guoSparseDNNsImproved2018,jakubovitz2018improving,ross2018improving} often evaluated the robustness of specific attack types without discussing the generalizability to different attacks.

    \textbf{Discussion of alternative specifications.} Fairness is a complex concept with diverse cultural, legal, societal, and ethical understandings. 
    Given the multiple competing notions of fairness and many possibilities of fairness metrics, it is important that authors justify their choices or at least acknowledge these choices. For example, \citet{roh2021Sample} state that their method is limited to a specific group of fairness measures (i.e., equalized odds and demographic parity disparity), and that one needs to choose a fairness measure in light of the underlying social context. 
    % In other words, the chosen evaluation metrics typically can not fully capture the desired aspects of model performance or evaluation may involve subjective measures that introduce variability in the results. 
    In contrast, other papers \cite{coston2020Counterfactual,komiyamaNonconvexOptimizationRegression2018, madrasPredictResponsiblyImproving2018} do not justify the choice of fairness metric or acknowledge alternative formulations.

    \textbf{Making assumptions explicit.} Some papers restrict the scope of the specification by identifying the assumptions behind it and the consequences of breaking some of them.
    For example, \citet{croce2021RobustBench} state that results on RobustBench, the proposed robustness benchmark, may not generalize well to real-world deployment if the data comes from a new domain or if novel adversarial attacks are used. 
    In contrast, \citet{gan2019Improving} train a system to generate paraphrases to test and improve the robustness of question-answering systems under the---implicit and not discussed---assumption that the system will generalize from system-generated paraphrases to real-world cases, which might not be the case.

    % \textbf{} MEMO \cite{zhang2022MEMO} introduces additional augmentation and adaptation procedures than standard model inference to improve model performance. However, a notable consequence is the increased demand for computational resources to process the augmented data which is addressed as a limitation in the paper. Similarly, AUGMIX \cite{hendrycks2020augmix} proposes stochasticity and diverse augmentations, which also increase computational and memory requirements during training, but these limitations are not mentioned in the paper.

\subsection{Analysis of recommendations}
\label{sec:analysis_of_recommendations}
We extracted recommendations regarding the proposed specification metric or optimization strategy from the analyzed papers.
We consider recommendations to be passages offering guidelines on integrating the specification into the system development process or how to interpret the metric alongside the task metric and other specifications when considering practical implications.
Recommendations can prevent misuse of the proposed specification or optimization strategy, such as applying a technique in the wrong context or falsely taking good performance on a specification to guarantee general system quality.

We categorize the extracted recommendations into the following types:

\textbf{Vague.} 
Vague recommendations provide high-level suggestions but do not define concrete measures that should be taken to enforce them.
Some works that propose specifications mention how they should \emph{supplement} standard evaluation but not \emph{substitute} it~\cite{wu2020INT,naik2018stress}.
While it is valuable to restrict the scope of the metric in that way, such guidelines are not directly actionable as they leave out the matter of \textit{how} the specification metric can supplement standard evaluation.

\textbf{Delegating.}
Delegating recommendations also provide abstract guidelines, but they defer the definition and execution of the guidelines to other actors.
An example is deferring results interpretations to domain experts~\cite{black2020FlipTest}.
Deferring decisions to the actors in the best position to make them is surely a good idea, but such recommendations often do not explicitly describe which factors the experts to which interpretation is delegated should consider when dealing with the specification.

\textbf{Debugging.} Some papers recommend that specifications be used for debugging, i.e., finding and fixing errors. For example, \citet{ribeiro2020beyond} proposes comprehensive and structured test suites to identify NLP systems' failure cases (e.g., robustness to typos).
\citet{rottger2021hatecheck} recommends fixing errors by sampling or constructing additional training examples resembling failed test cases.
% Similarly, HATECHECK \cite{rottger2020hatecheck} recommends users to resemble test cases from inaccurate model functional tests, for instance, sample or construct additional training cases to resemble test cases from functional tests that their model was inaccurate on, bearing in mind that this additional data might introduce other unforeseen biases. 

% "This data should not be used to train a language model on a language modeling, or masked language modeling, objective. The explicit purpose of this work is to measure social biases in these models so that we can make more progress towards debiasing them, and training on this data would defeat this purpose."
\textbf{(Not) additional data.} Some works that propose example-based specifications explicitly state how the data should or should not be used for system development.
Researchers may recommend that the data should be used to optimize the specification.
\citet{kirk2022Hatemojia} create two datasets for hate speech detection of emoji-containing texts: one for testing and one for optimization.
Conversely, \emph{\textbf{not} additional data} recommendations state that the data is purely for testing and should not be used for training. For example,  \citet{nangia2020crows} state that CrowS-pairs, the proposed dataset, should be used to measure social biases, not for debiasing systems, stating that debiasing a system in a way that generalizes is challenging and may require larger datasets.

\textbf{Concrete.} In contrast to vague recommendations, concrete recommendations provide comprehensive and detailed recommendations of how to use the specification in the development cycle.
%their proposed work, which can contain several categories of recommendations such as the ones mentioned above. 
E.g.,  \citet{pfohl2022Net} compare several methods to improve the fairness of predictive systems in healthcare, considering both system performance and fairness measures.
They use this empirical analysis to recommend a specific fairness criterion (subpopulation-specific calibration), describing how to apply it for system development (prioritizing systems based on validation-set calibrations and setting subpopulation-specific decision thresholds), and what other factors should be considered (transparency, participation of stakeholders in the decision processes, and reasoning about the potential impact of system-informed decisions).

\subsection{Case studies}
This section presents three representative papers that illustrate our survey questions and aspects of specification overfitting.

\paragraph{\hatecheck: Functional Tests for Hate Speech Detection Models~\cite{rottger2021hatecheck}}
The paper introduces \hatecheck, a test suite for hate speech detection.
\hatecheck covers 29 distinct functionalities that examine distinct expressions of hate (e.g., implicit derogation and hate expressed using slur) and contrastive non-hate (e.g., denouncement of hate that quotes it, or abuse targeted at objects).

We view each functionality as a distinct specification corresponding to an underlying system capability.
The specifications are example-based: each functionality is assessed through a set of test cases that are either handcrafted (h) or generated through templates (pat).
The specification metric is the accuracy computed on the functionality's test cases.

Though the paper does not experiment with specification optimization strategies, it still contrasts specification performance with general task correctness using two standard hate speech datasets~\cite{founta2018large,davidson2017automated}.
As the suite comprises multiple functionalities, multiple specification values are reported.

\citet{rottger2021hatecheck} discuss \hatecheck's scope and limitations in a dedicated section, highlighting how \hatecheck has limited coverage.
That is, good performance on \hatecheck only reveals the absence of weakness for the tested cases, not generalizable strengths.
Notably, the \hatecheck benchmark does not produce insight into phenomena that are not covered (e.g., involving other protected groups, languages, and combinations of functionalities).

The impact statement section summarizes \hatecheck's scope:
  \begin{quote}
   ``\hatecheck's intended use is as an evaluative tool for hate speech detection models, providing structured and targeted diagnostic insights into model functionalities. [...]
   Researchers might overextend claims about the functionalities of their models based on their test performance, which we would consider a misuse of \hatecheck'' \cite[p.\ 50]{rottger2021hatecheck}.
  \end{quote}

In addition to stating what \hatecheck should \emph{not} be used for, it points out \emph{how}  \hatecheck can aid system development:
\begin{quote}
  ``If poor model performance does stem from biased training data, models could be improved through targeted data augmentation \cite{gardner2020evaluating}. 
  \hatecheck users could, for instance, sample or construct additional training cases to resemble test cases from functional tests that their model was inaccurate on, bearing in mind that this additional data might introduce other unforeseen biases.
  The models we tested would likely benefit from training on additional cases of negated hate, reclaimed slurs and counter speech" \cite[p.\ 48]{rottger2021hatecheck}.
\end{quote}

This recommendation contextualizes the specifications vis-a-vis system development (suggests augmenting training data with cases similar to the suite) and points out a possible pitfall---introducing unforeseen biases.

\paragraph{Benchmarking Neural Network Robustness to Common Corruptions and Perturbations \cite{hendrycks2018Benchmarking}}

The paper introduces a benchmark that measures the robustness of image classifiers. 
% This is different from previous work that has focused on the challenges of robustness to adversarial examples \cite{carlini2017adversarial}.
Specifically, it proposes two datasets -- IMAGENET-C, obtained by modifying images from ImageNet \cite{deng2009imagenet} using a set of 75 algorithmically generated corruptions) and IMAGENET-P, which includes sequences where an image is gradually perturbed with similar corruptions from IMAGENET-C.
To validate their datasets, the authors show that there is wide room for improvement on IMAGENET-C by evaluating the performance of several deep learning systems. Additionally, they introduced three methods and architectures that improve corruption robustness. For IMAGENET-P, they propose metrics to measure the stability of the network’s predictions on the perturbed images. 

The authors state the goal of IMAGETNET-C in the following manner:

\begin{quote}
    ``We hope that this will serve as a general dataset for benchmarking robustness to image corruptions and prevent methodological problems such as moving goal posts and result cherry picking."
\end{quote}

Moreover, they recommend future work to use this benchmark because:

\begin{quote}
    ``By defining and benchmarking perturbation and corruption robustness, we facilitate research that can be overcome by future networks which do not rely on spurious correlations or cues inessential to the object’s class."
\end{quote}

The provided recommendation asserts the importance of measuring perturbation and corruption robustness, but how to act on the insights provided by the benchmark is not discussed, i.e., how to improve system corruption robustness, overall accuracy, and other notions of robustness, such as adversarial robustness.
% Therefore, evaluating methods only on this benchmark provides limited insight into the robustness against adversarial perturbations. 
Later work \cite{schneider2020Improving} illustrates how focusing on one type of robustness may provide limited insight into other types:
 \begin{quote}

 ``We here argue that popular benchmarks to measure model robustness against common corruptions (like ImageNet-C) underestimate model robustness in many (but not all) application scenarios."\\
 ``So far, popular image corruption benchmarks like ImageNet-C focus only on ad hoc scenarios in which the tested model has zero prior knowledge about the corruptions it encounters during test time, even if it encounters the same corruption multiple times."

\end{quote}

\paragraph{Net benefit, calibration, threshold selection, and training objectives for algorithmic fairness in healthcare \cite{pfohl2022Net}}

This work compares the estimated \emph{net utility} of predictive systems in healthcare. Specifically, the authors train predictive systems that output a continuous-valued risk score (risk of atherosclerotic cardiovascular disease), which serves as the sole basis for a hypothetical clinical intervention (statin initiation based on decision thresholds).
The utility itself is estimated by a secondary system, which parametrizes the relative value of the harms and benefits of the (hypothetical) clinical intervention according to clinical data.

The article reports the overall net utility for the entire patient pool in the data set for different predictive systems, as well as the utility for different subgroups of patients according to sex, racial, and ethnic categories, and the presence of type 2 and type 1 diabetes, rheumatoid arthritis, and chronic kidney disease. Moreover, in addition to the utility itself, an in-depth analysis of the results is reported, including the measurement of equalized odds \cite{hardt2016equality} as a metric to measure fairness across intersectional subgroups (combining race, ethnicity, and sex). Different methods for improving fairness are compared, specifically comparing \emph{in-processing} approaches \cite{pfohl2022comparison} that aim at producing a fair system penalizing worst-group performance during training with \emph{post-processing} approaches that learn the predictive system in an unconstrained manner (unpenalized empirical risk minimization, ERM) and calibrate the decision thresholds to improve fairness on the resulting system.

In our analysis of this article, we view the overall net utility for the entire patient pool as the \emph{task metric} and equalized odds as the \emph{specification metric}. We categorize the \emph{in-processing} and \emph{post-processing} strategies as \emph{direct} attempts to optimize fairness.

The authors of \cite{pfohl2022Net}, in contrast to most other papers in our collection, have clear recommendations on how to use the specification metric during the development process, advising against \emph{in-processing} methods and for threshold calibration as a \emph{post-processing} step:

 \begin{quote}

 ``[...] approaches that incorporate fairness considerations into the model training objective typically do not improve model performance or confer greater net benefit"\\
 ``[...] we argue for focusing model development efforts on developing calibrated models that predict outcomes well for all patient populations while emphasizing that such efforts are complementary to transparent reporting, participatory design, and reasoning about the impact of model-informed interventions in context. "\\
 ``[...] results indicate that models derived from unpenalized ERM should not necessarily be assumed to be well-calibrated in practice, further highlighting the importance of model development, selection, and post-processing strategies that aims to identify the best-fitting, well-calibrated model for each subgroup."

\end{quote}

\section{Conclusion}
\label{sec:conclusion}
In this article, we discussed the problem of specification overfitting---improving specified metrics to the detriment of the underlying goal or other metrics.
We analyzed recent impactful papers from diverse AI fields to identify if and how works that propose specification metrics or improvement methods consider specification overfitting.
We have found that specification overfitting is often implicitly addressed, with most papers reporting the main task metric or more than one specification metric.
However, papers rarely discuss the role of specifications in the system development process, leaving out questions such as how to integrate several (possibly competing) metrics and the assumptions underlying the formulation of a metric.
Works that discuss these questions frequently do it in a vague way or leave decision-making to users or domain experts without providing guidelines on how to make such decisions.

Given that the currently developing legislative frameworks use broad terms for the requirements for AI systems, AI providers wanting to achieve legal compliance need to rely on standardized specification metrics set by standardization organizations. 
Therefore, specifications gain enormous importance in the legal framework and should be carefully evaluated, especially given the specification overfitting issues discussed in this paper. 
If these are not duly considered on the regulatory and standard-setting level, citizens may not be sufficiently protected from potential harm.

We recommend metric proposers be explicit about how the metric differs from the ideal property it intends to measure.
Given that the metric may disagree with other measures of system quality, we also recommend that they provide guidelines or recommendations on making decisions on system selection.
We recommend that peer reviewers reward papers with clear delimitations of the scope of a specification metric and that mentioning such limitations should not be seen as a weakness.
Method proposers should rigorously measure the impact of the method on other metrics, including the task metric.
One way to do so is by defining evaluation scenarios that are robust to specification overfitting, such as using controlled splits that hold out metrics.
Practitioners, regulators, and standard-setting bodies should be aware of the misincentives that can arise from using a narrow set of metrics for evaluation when these same metrics can be a target in optimization and system selection.

\backmatter

% \bmhead{Supplementary information}

\bmhead{Acknowledgments}

This research has been funded by the Vienna Science and Technology Fund (WWTF) [10.47379/VRG19008] ``Knowledge-infused Deep Learning for Natural Language Processing''.

\bmhead{Author contributions}
BR conceptualized the study, contributed to the study design, and analyzed papers for the survey.
PHLA contributed to the study design and collected, filtered, and analyzed papers for the survey.
YX contributed to the study design and analyzed papers for the survey.
SK and CK bridged the technical aspects of the study to the legal framework and regulatory practices.
All authors contributed to the paper writing and revising.

\section*{Declarations}

\bmhead{Competing interests} The authors have no competing interests to declare that are relevant to the content of this article.

\bibliography{refs}% common bib file

\begin{thebibliography}{170}
\providecommand{\natexlab}[1]{#1}
\providecommand{\url}[1]{{#1}}
\providecommand{\urlprefix}{URL }
\providecommand{\doi}[1]{\url{https://doi.org/#1}}
\providecommand{\eprint}[2][]{\url{#2}}
 \bibcommenthead

\bibitem[{Ace{\~n}a et~al(2022)Ace{\~n}a, {Mart{\'i}n de Diego},
  R.~Fern{\'a}ndez, and M.~Moguerza}]{acena2022Minimally}
Ace{\~n}a V, {Mart{\'i}n de Diego} I, R.~Fern{\'a}ndez R, et~al (2022)
  Minimally overfitted learners: {{A}} general framework for ensemble learning.
  Knowledge-Based Systems 254:109669. \doi{10.1016/j.knosys.2022.109669}

\bibitem[{Angwin et~al(2016)Angwin, Jeff~Larson, and
  Kirchner}]{angwin2016machine}
Angwin J, Jeff~Larson SM, Kirchner L (2016) Machine {{Bias}}. Pro Publica
  \urlprefix\url{https://www.propublica.org/article/machine-bias-risk-assessments-in-criminal-sentencing}

\bibitem[{Arulkumaran et~al(2017)Arulkumaran, Deisenroth, Brundage, and
  Bharath}]{arulkamaran2017deep}
Arulkumaran K, Deisenroth MP, Brundage M, et~al (2017) Deep reinforcement
  learning: A brief survey. IEEE Signal Processing Magazine 34(6):26--38.
  \doi{10.1109/MSP.2017.2743240}

\bibitem[{Bahdanau et~al(2015)Bahdanau, Cho, and Bengio}]{bahdanu2015neural}
Bahdanau D, Cho K, Bengio Y (2015) Neural machine translation by jointly
  learning to align and translate. In: Bengio Y, LeCun Y (eds) 3rd
  International Conference on Learning Representations, {ICLR} 2015, San Diego,
  CA, USA, May 7-9, 2015, Conference Track Proceedings,
  \urlprefix\url{http://arxiv.org/abs/1409.0473}

\bibitem[{Barocas and Selbst(2016)}]{barocas2016big}
Barocas S, Selbst AD (2016) Big data's disparate impact. Calif L Rev 104:671

\bibitem[{Barocas et~al(2019)Barocas, Hardt, and
  Narayanan}]{barocas2019fairness}
Barocas S, Hardt M, Narayanan A (2019) Fairness and Machine Learning:
  Limitations and Opportunities. fairmlbook.org,
  \url{http://www.fairmlbook.org}

\bibitem[{Bartolo et~al(2021)Bartolo, Thrush, Jia, Riedel, Stenetorp, and
  Kiela}]{bartolo2021Improving}
Bartolo M, Thrush T, Jia R, et~al (2021) Improving {{Question Answering Model
  Robustness}} with {{Synthetic Adversarial Data Generation}}. In: Moens MF,
  Huang X, Specia L, et~al (eds) Proceedings of the 2021 {{Conference}} on
  {{Empirical Methods}} in {{Natural Language Processing}}. {Association for
  Computational Linguistics}, {Online and Punta Cana, Dominican Republic}, pp
  8830--8848, \doi{10.18653/v1/2021.emnlp-main.696}

\bibitem[{Ben-David et~al(2010)Ben-David, Blitzer, Crammer, Kulesza, Pereira,
  and Vaughan}]{bendavid2010theory}
Ben-David S, Blitzer J, Crammer K, et~al (2010) A theory of learning from
  different domains. Machine Learning 79(1):151--175.
  \doi{10.1007/s10994-009-5152-4}

\bibitem[{Bhatore et~al(2020)Bhatore, Mohan, and Reddy}]{bhatore2020machine}
Bhatore S, Mohan L, Reddy YR (2020) Machine learning techniques for credit risk
  evaluation: a systematic literature review. Journal of Banking and Financial
  Technology 4(1):111--138. \doi{10.1007/s42786-020-00020-3}

\bibitem[{Black et~al(2020)Black, Yeom, and Fredrikson}]{black2020FlipTest}
Black E, Yeom S, Fredrikson M (2020) {{FlipTest}}: Fairness testing via optimal
  transport. In: Proceedings of the 2020 {{Conference}} on {{Fairness}},
  {{Accountability}}, and {{Transparency}}. {ACM}, {Barcelona Spain}, pp
  111--121, \doi{10.1145/3351095.3372845}

\bibitem[{Bleher and Braun(2023)}]{bleher2023Reflections}
Bleher H, Braun M (2023) Reflections on {{Putting AI Ethics}} into
  {{Practice}}: {{How Three AI Ethics Approaches Conceptualize Theory}} and
  {{Practice}}. Science and Engineering Ethics 29(3):21.
  \doi{10.1007/s11948-023-00443-3}

\bibitem[{Borisov et~al(2022)Borisov, Leemann, Seßler, Haug, Pawelczyk, and
  Kasneci}]{borisov2022deep}
Borisov V, Leemann T, Seßler K, et~al (2022) Deep neural networks and tabular
  data: A survey. IEEE Transactions on Neural Networks and Learning Systems pp
  1--21. \doi{10.1109/TNNLS.2022.3229161}

\bibitem[{Bronstein et~al(2017)Bronstein, Bruna, LeCun, Szlam, and
  Vandergheynst}]{bronstein2017Geometric}
Bronstein MM, Bruna J, LeCun Y, et~al (2017) Geometric {{Deep Learning}}:
  {{Going}} beyond {{Euclidean}} data. IEEE Signal Processing Magazine
  34(4):18--42. \doi{10.1109/MSP.2017.2693418}

\bibitem[{Buffelli et~al(2022)Buffelli, Li{\'o}, and
  Vandin}]{buffelli2022SizeShiftReg}
Buffelli D, Li{\'o} P, Vandin F (2022) {{SizeShiftReg}}: A {{Regularization
  Method}} for {{Improving Size-Generalization}} in {{Graph Neural Networks}}.
  Advances in Neural Information Processing Systems 35:31871--31885.
  \urlprefix\url{https://proceedings.neurips.cc/paper_files/paper/2022/file/ceeb3fa5be458f08fbb12a5bb783aac8-Paper-Conference.pdf}

\bibitem[{Buolamwini and Gebru(2018)}]{buolamwini2018gender}
Buolamwini J, Gebru T (2018) Gender shades: Intersectional accuracy disparities
  in commercial gender classification. In: Friedler SA, Wilson C (eds)
  Proceedings of the 1st Conference on Fairness, Accountability and
  Transparency, Proceedings of Machine Learning Research, vol~81. PMLR, pp
  77--91, \urlprefix\url{https://proceedings.mlr.press/v81/buolamwini18a.html}

\bibitem[{Chen et~al(2019)Chen, Kallus, Mao, Svacha, and
  Udell}]{chen2019Fairness}
Chen J, Kallus N, Mao X, et~al (2019) Fairness {{Under Unawareness}}:
  {{Assessing Disparity When Protected Class Is Unobserved}}. In: Proceedings
  of the {{Conference}} on {{Fairness}}, {{Accountability}}, and
  {{Transparency}}. {Association for Computing Machinery}, {New York, NY, USA},
  {{FAT}}* '19, pp 339--348, \doi{10.1145/3287560.3287594}

\bibitem[{Chen et~al(2022)Chen, Zhou, Bian, Xie, Wu, Zhang, Kaili, Yang, Zhao,
  Han, and Cheng}]{chen2022Pareto}
Chen Y, Zhou K, Bian Y, et~al (2022) Pareto {{Invariant Risk Minimization}}:
  {{Towards Mitigating}} the {{Optimization Dilemma}} in {{Out-of-Distribution
  Generalization}}. In: The {{Eleventh International Conference}} on {{Learning
  Representations}}

\bibitem[{Cheng et~al(2019)Cheng, Wei, and Hsieh}]{cheng2019Evaluating}
Cheng M, Wei W, Hsieh CJ (2019) Evaluating and {{Enhancing}} the {{Robustness}}
  of {{Dialogue Systems}}: {{A Case Study}} on a {{Negotiation Agent}}. In:
  Burstein J, Doran C, Solorio T (eds) Proceedings of the 2019 {{Conference}}
  of the {{North American Chapter}} of the {{Association}} for {{Computational
  Linguistics}}: {{Human Language Technologies}}, {{Volume}} 1 ({{Long}} and
  {{Short Papers}}). {Association for Computational Linguistics}, {Minneapolis,
  Minnesota}, pp 3325--3335, \doi{10.18653/v1/N19-1336}

\bibitem[{Cheng et~al(2020)Cheng, Yi, Chen, Zhang, and
  Hsieh}]{cheng2020Seq2Sick}
Cheng M, Yi J, Chen PY, et~al (2020) {{Seq2Sick}}: {{Evaluating}} the
  {{Robustness}} of {{Sequence-to-Sequence Models}} with {{Adversarial
  Examples}}. Proceedings of the AAAI Conference on Artificial Intelligence
  34(04):3601--3608. \doi{10.1609/aaai.v34i04.5767}

\bibitem[{Cheng et~al(2022)Cheng, Lei, Chen, Dhillon, and Hsieh}]{cheng2022cat}
Cheng M, Lei Q, Chen PY, et~al (2022) {{CAT}}: {{Customized Adversarial
  Training}} for {{Improved Robustness}}. In: Raedt LD (ed) Proceedings of the
  {{Thirty-First International Joint Conference}} on {{Artificial
  Intelligence}}, {{IJCAI}} 2022, {{Vienna}}, {{Austria}}, 23-29 {{July}} 2022.
  {ijcai.org}, pp 673--679, \doi{10.24963/IJCAI.2022/95}

\bibitem[{Clark et~al(2020)Clark, Luong, Le, and Manning}]{clark2020ELECTRA}
Clark K, Luong MT, Le QV, et~al (2020) {ELECTRA: Pre-training Text Encoders as
  Discriminators Rather Than Generators}. In: International Conference on
  Learning Representations,
  \urlprefix\url{https://openreview.net/forum?id=r1xMH1BtvB}

\bibitem[{Clarysse et~al(2022)Clarysse, H{\"o}rrmann, and
  Yang}]{clarysse2022Why}
Clarysse J, H{\"o}rrmann J, Yang F (2022) Why adversarial training can hurt
  robust accuracy. In: The {{Eleventh International Conference}} on {{Learning
  Representations}}

\bibitem[{Coston et~al(2020)Coston, Mishler, Kennedy, and
  Chouldechova}]{coston2020Counterfactual}
Coston A, Mishler A, Kennedy EH, et~al (2020) Counterfactual risk assessments,
  evaluation, and fairness. In: Proceedings of the 2020 {{Conference}} on
  {{Fairness}}, {{Accountability}}, and {{Transparency}}. {Association for
  Computing Machinery}, {New York, NY, USA}, {{FAT}}* '20, pp 582--593,
  \doi{10.1145/3351095.3372851}, \eprint{1909.00066}

\bibitem[{Cotter et~al(2019)Cotter, Gupta, Jiang, Srebro, Sridharan, Wang,
  Woodworth, and You}]{cotter2019Training}
Cotter A, Gupta M, Jiang H, et~al (2019) Training {{Well-Generalizing
  Classifiers}} for {{Fairness Metrics}} and {{Other Data-Dependent
  Constraints}}. In: Proceedings of the 36th {{International Conference}} on
  {{Machine Learning}}. {PMLR}, pp 1397--1405

\bibitem[{Croce et~al(2021)Croce, Andriushchenko, Sehwag, Debenedetti,
  Flammarion, Chiang, Mittal, and Hein}]{croce2021RobustBench}
Croce F, Andriushchenko M, Sehwag V, et~al (2021) {{RobustBench}}: A
  standardized adversarial robustness benchmark. In: Thirty-Fifth
  {{Conference}} on {{Neural Information Processing Systems Datasets}} and
  {{Benchmarks Track}} ({{Round}} 2), \eprint{2010.09670}

\bibitem[{D'Amour et~al(2022)D'Amour, Heller, Moldovan, Adlam, Alipanahi,
  Beutel, Chen, Deaton, Eisenstein, Hoffman et~al}]{d2022underspecification}
D'Amour A, Heller K, Moldovan D, et~al (2022) Underspecification presents
  challenges for credibility in modern machine learning. The Journal of Machine
  Learning Research 23(1):10237--10297

\bibitem[{Dapello et~al(2022)Dapello, Kar, Schrimpf, Geary, Ferguson, Cox, and
  DiCarlo}]{dapello2022Aligning}
Dapello J, Kar K, Schrimpf M, et~al (2022) Aligning {{Model}} and {{Macaque
  Inferior Temporal Cortex Representations Improves Model-to-Human Behavioral
  Alignment}} and {{Adversarial Robustness}}. In: The {{Eleventh International
  Conference}} on {{Learning Representations}}. Cold Spring Harbor Laboratory,
  pp 2022--07

\bibitem[{Davidson et~al(2017)Davidson, Warmsley, Macy, and
  Weber}]{davidson2017automated}
Davidson T, Warmsley D, Macy M, et~al (2017) Automated hate speech detection
  and the problem of offensive language. Proceedings of the International AAAI
  Conference on Web and Social Media 11(1):512--515.
  \urlprefix\url{https://ojs.aaai.org/index.php/ICWSM/article/view/14955}

\bibitem[{Deng et~al(2009)Deng, Dong, Socher, Li, Li, and
  Fei-Fei}]{deng2009imagenet}
Deng J, Dong W, Socher R, et~al (2009) Imagenet: A large-scale hierarchical
  image database. In: 2009 IEEE conference on computer vision and pattern
  recognition, Ieee, pp 248--255

\bibitem[{Deng et~al(2023)Deng, Zhang, Zhang, Ye, Coley, Su, and
  Zou}]{deng2023FIFA}
Deng Z, Zhang J, Zhang L, et~al (2023) {FIFA:} making fairness more
  generalizable in classifiers trained on imbalanced data. In: The Eleventh
  International Conference on Learning Representations, {ICLR} 2023, Kigali,
  Rwanda, May 1-5, 2023. OpenReview.net,
  \urlprefix\url{https://openreview.net/pdf?id=zVrw4OH1Lch}

\bibitem[{Devlin et~al(2019)Devlin, Chang, Lee, and Toutanova}]{devlin2019bert}
Devlin J, Chang MW, Lee K, et~al (2019) {BERT}: Pre-training of deep
  bidirectional transformers for language understanding. In: Proceedings of the
  2019 Conference of the North {A}merican Chapter of the Association for
  Computational Linguistics: Human Language Technologies, Volume 1 (Long and
  Short Papers). Association for Computational Linguistics, Minneapolis,
  Minnesota, pp 4171--4186,
  \urlprefix\url{https://www.aclweb.org/anthology/N19-1423}

\bibitem[{Dosovitskiy et~al(2021)Dosovitskiy, Beyer, Kolesnikov, Weissenborn,
  Zhai, Unterthiner, Dehghani, Minderer, Heigold, Gelly, Uszkoreit, and
  Houlsby}]{kolesnikov2021image}
Dosovitskiy A, Beyer L, Kolesnikov A, et~al (2021) An image is worth 16x16
  words: Transformers for image recognition at scale. In: International
  Conference on Learning Representations,
  \urlprefix\url{https://openreview.net/forum?id=YicbFdNTTy}

\bibitem[{Elkahky et~al(2018)Elkahky, Webster, Andor, and
  Pitler}]{elkahkyChallengeSetMethods2018}
Elkahky A, Webster K, Andor D, et~al (2018) A {{Challenge Set}} and {{Methods}}
  for {{Noun-Verb Ambiguity}}. In: Proceedings of the 2018 {{Conference}} on
  {{Empirical Methods}} in {{Natural Language Processing}}. {Association for
  Computational Linguistics}, {Brussels, Belgium}, pp 2562--2572,
  \doi{10.18653/v1/D18-1277}

\bibitem[{Esteva et~al(2021)Esteva, Chou, Yeung, Naik, Madani, Mottaghi, Liu,
  Topol, Dean, and Socher}]{esteva2021deep}
Esteva A, Chou K, Yeung S, et~al (2021) Deep learning-enabled medical computer
  vision. npj Digital Medicine 4(1):5. \doi{10.1038/s41746-020-00376-2}

\bibitem[{{European Parliament and Council of the European
  Union}(2012)}]{EU2012standardization}
{European Parliament and Council of the European Union} (2012) {Regulation (EU)
  No 1025/2012 of the European Parliament and of the Council of 25 October 2012
  on European standardisation}.
  \urlprefix\url{https://eur-lex.europa.eu/eli/reg/2012/1025/oj}

\bibitem[{{European Parliament and Council of the European
  Union}(2022)}]{EU2022proposal}
{European Parliament and Council of the European Union} (2022) {Proposal for a
  Directive of the European Parliament and of the Council on adapting
  non-contractual civil liability rules to artificial intelligence (AI
  Liability Directive)}.
  \urlprefix\url{https://eur-lex.europa.eu/legal-content/EN/TXT/?uri=CELEX%3A52022PC0496}

\bibitem[{{European Parliament and Council of the European
  Union}(2024)}]{EU2024AIAct}
{European Parliament and Council of the European Union} (2024) {Regulation of
  the European Parliament and of the Council laying down harmonised rules on
  artificial intelligence (Artificial intelligence Act)}

\bibitem[{Fan et~al(2019)Fan, Ma, Li, He, Zhao, Tang, and Yin}]{fan2019graph}
Fan W, Ma Y, Li Q, et~al (2019) {Graph Neural Networks for Social
  Recommendation}. In: The World Wide Web Conference. Association for Computing
  Machinery, New York, NY, USA, WWW '19, p 417–426,
  \doi{10.1145/3308558.3313488}

\bibitem[{Fatemi et~al(2023)Fatemi, Xing, Liu, and Xiong}]{fatemi2023Improving}
Fatemi Z, Xing C, Liu W, et~al (2023) Improving {{Gender Fairness}} of
  {{Pre-Trained Language Models}} without {{Catastrophic Forgetting}}. In:
  Rogers A, {Boyd-Graber} J, Okazaki N (eds) Proceedings of the 61st {{Annual
  Meeting}} of the {{Association}} for {{Computational Linguistics}}
  ({{Volume}} 2: {{Short Papers}}). {Association for Computational
  Linguistics}, {Toronto, Canada}, pp 1249--1262,
  \doi{10.18653/v1/2023.acl-short.108}

\bibitem[{Fjeld et~al(2020)Fjeld, Achten, Hilligoss, Nagy, and
  Srikumar}]{fjeld2020principled}
Fjeld J, Achten N, Hilligoss H, et~al (2020) {Principled artificial
  intelligence: Mapping consensus in ethical and rights-based approaches to
  principles for AI}. Berkman Klein Center Research Publication 2020-1

\bibitem[{Founta et~al(2018)Founta, Djouvas, Chatzakou, Leontiadis, Blackburn,
  Stringhini, Vakali, Sirivianos, and Kourtellis}]{founta2018large}
Founta A, Djouvas C, Chatzakou D, et~al (2018) {Large Scale Crowdsourcing and
  Characterization of Twitter Abusive Behavior}. Proceedings of the
  International AAAI Conference on Web and Social Media 12(1).
  \urlprefix\url{https://ojs.aaai.org/index.php/ICWSM/article/view/14991}

\bibitem[{Friedler et~al(2021)Friedler, Scheidegger, and
  Venkatasubramanian}]{friedler2021impossibility}
Friedler SA, Scheidegger C, Venkatasubramanian S (2021) The (im)possibility of
  fairness: different value systems require different mechanisms for fair
  decision making. Commun ACM 64(4):136–143. \doi{10.1145/3433949}

\bibitem[{Fukushima(1980)}]{fukushima1980neocognitron}
Fukushima K (1980) Neocognitron: A self-organizing neural network model for a
  mechanism of pattern recognition unaffected by shift in position. Biological
  Cybernetics 36(4):193--202. \doi{10.1007/BF00344251}

\bibitem[{Gan and Ng(2019)}]{gan2019Improving}
Gan WC, Ng HT (2019) Improving the {{Robustness}} of {{Question Answering
  Systems}} to {{Question Paraphrasing}}. In: Proceedings of the 57th {{Annual
  Meeting}} of the {{Association}} for {{Computational Linguistics}}.
  {Association for Computational Linguistics}, {Florence, Italy}, pp
  6065--6075, \doi{10.18653/v1/P19-1610}

\bibitem[{Gardner et~al(2020)Gardner, Artzi, Basmov, Berant, Bogin, Chen,
  Dasigi, Dua, Elazar, Gottumukkala, Gupta, Hajishirzi, Ilharco, Khashabi, Lin,
  Liu, Liu, Mulcaire, Ning, Singh, Smith, Subramanian, Tsarfaty, Wallace,
  Zhang, and Zhou}]{gardner2020evaluating}
Gardner M, Artzi Y, Basmov V, et~al (2020) Evaluating models{'} local decision
  boundaries via contrast sets. In: Findings of the Association for
  Computational Linguistics: EMNLP 2020. Association for Computational
  Linguistics, Online, pp 1307--1323,
  \doi{10.18653/v1/2020.findings-emnlp.117},
  \urlprefix\url{https://aclanthology.org/2020.findings-emnlp.117}

\bibitem[{Geirhos et~al(2018)Geirhos, Rubisch, Michaelis, Bethge, Wichmann, and
  Brendel}]{geirhos2018ImageNettrained}
Geirhos R, Rubisch P, Michaelis C, et~al (2018) {{ImageNet-trained CNNs}} are
  biased towards texture; increasing shape bias improves accuracy and
  robustness. In: International {{Conference}} on {{Learning Representations}}

\bibitem[{Goodfellow et~al(2015)Goodfellow, Shlens, and
  Szegedy}]{goodfellow2015explaining}
Goodfellow I, Shlens J, Szegedy C (2015) Explaining and harnessing adversarial
  examples. In: International Confserence on Learning Representations,
  \urlprefix\url{http://arxiv.org/abs/1412.6572}

\bibitem[{Gowal et~al(2021)Gowal, Rebuffi, Wiles, Stimberg, Calian, and
  Mann}]{gowal2021Improving}
Gowal S, Rebuffi SA, Wiles O, et~al (2021) Improving {{Robustness}} using
  {{Generated Data}}. In: Advances in {{Neural Information Processing
  Systems}}, vol~34. {Curran Associates, Inc.}, pp 4218--4233

\bibitem[{Guo et~al(2022)Guo, Chen, Hao, Yin, Yu, and
  Li}]{guo2022Comprehensive}
Guo J, Chen Y, Hao Y, et~al (2022) Towards {{Comprehensive Testing}} on the
  {{Robustness}} of {{Cooperative Multi-agent Reinforcement Learning}}. In:
  {{IEEE}}/{{CVF Conference}} on {{Computer Vision}} and {{Pattern Recognition
  Workshops}}, {{CVPR Workshops}} 2022, {{New Orleans}}, {{LA}}, {{USA}},
  {{June}} 19-20, 2022. {IEEE}, pp 114--121,
  \doi{10.1109/CVPRW56347.2022.00022}

\bibitem[{Guo et~al(2018)Guo, Zhang, Zhang, and
  Chen}]{guoSparseDNNsImproved2018}
Guo Y, Zhang C, Zhang C, et~al (2018) Sparse {{DNNs}} with {{Improved
  Adversarial Robustness}}. In: Advances in {{Neural Information Processing
  Systems}}, vol~31. {Curran Associates, Inc.}

\bibitem[{Hagendorff(2020)}]{hagendorff2020ethics}
Hagendorff T (2020) {The Ethics of AI Ethics: An Evaluation of Guidelines}.
  Minds and Machines 30(1):99--120. \doi{10.1007/s11023-020-09517-8}

\bibitem[{Han et~al(2023)Han, Wang, Chen, Chen, Guo, Liu, Tang, Xiao, Xu, Xu,
  Yang, Zhang, and Tao}]{han2023survey}
Han K, Wang Y, Chen H, et~al (2023) {A Survey on Vision Transformer}. IEEE
  Transactions on Pattern Analysis and Machine Intelligence 45(1):87--110.
  \doi{10.1109/TPAMI.2022.3152247}

\bibitem[{Hardt et~al(2016)Hardt, Price, and Srebro}]{hardt2016equality}
Hardt M, Price E, Srebro N (2016) Equality of opportunity in supervised
  learning. Advances in neural information processing systems 29

\bibitem[{Havasi et~al(2020)Havasi, Jenatton, Fort, Liu, Snoek,
  Lakshminarayanan, Dai, and Tran}]{havasi2020Training}
Havasi M, Jenatton R, Fort S, et~al (2020) Training independent subnetworks for
  robust prediction. In: International {{Conference}} on {{Learning
  Representations}}

\bibitem[{He et~al(2016)He, Zhang, Ren, and Sun}]{he2016deep}
He K, Zhang X, Ren S, et~al (2016) Deep residual learning for image
  recognition. In: 2016 IEEE Conference on Computer Vision and Pattern
  Recognition (CVPR), pp 770--778, \doi{10.1109/CVPR.2016.90}

\bibitem[{He et~al(2005)He, Yan, Hu, Niyogi, and Zhang}]{he2005face}
He X, Yan S, Hu Y, et~al (2005) {Face recognition using Laplacianfaces}. IEEE
  Transactions on Pattern Analysis and Machine Intelligence 27(3):328--340.
  \doi{10.1109/TPAMI.2005.55}

\bibitem[{Hendrycks and Dietterich(2018)}]{hendrycks2018Benchmarking}
Hendrycks D, Dietterich T (2018) Benchmarking {{Neural Network Robustness}} to
  {{Common Corruptions}} and {{Perturbations}}. In: International
  {{Conference}} on {{Learning Representations}}

\bibitem[{Hendrycks et~al(2019{\natexlab{a}})Hendrycks, Mazeika, Kadavath, and
  Song}]{hendrycks2019Usinga}
Hendrycks D, Mazeika M, Kadavath S, et~al (2019{\natexlab{a}}) Using
  {{Self-Supervised Learning Can Improve Model Robustness}} and
  {{Uncertainty}}. In: Advances in {{Neural Information Processing Systems}},
  vol~32. {Curran Associates, Inc.}

\bibitem[{Hendrycks et~al(2019{\natexlab{b}})Hendrycks, Mu*, Cubuk, Zoph,
  Gilmer, and Lakshminarayanan}]{hendrycks*2019AugMix}
Hendrycks D, Mu* N, Cubuk ED, et~al (2019{\natexlab{b}}) {{AugMix}}: {{A Simple
  Data Processing Method}} to {{Improve Robustness}} and {{Uncertainty}}. In:
  International {{Conference}} on {{Learning Representations}}

\bibitem[{{High-Level Expert Group on AI}(2019)}]{HLEG2019guidelines}
{High-Level Expert Group on AI} (2019) {Ethics guidelines for trustworthy AI,
  High-Level Expert Group on AI}.
  \urlprefix\url{https://digital-strategy.ec.europa.eu/en/library/ethics-guidelines-trustworthy-ai}

\bibitem[{Hu et~al(2023)Hu, Yang, Chen, Li, Sima, Zhu, Chai, Du, Lin, Wang, Lu,
  Jia, Liu, Dai, Qiao, and Li}]{hu2023planning}
Hu Y, Yang J, Chen L, et~al (2023) Planning-oriented autonomous driving. In:
  Proceedings of the IEEE/CVF Conference on Computer Vision and Pattern
  Recognition

\bibitem[{Iniesta(2023)}]{iniesta2023human}
Iniesta R (2023) The human role to guarantee an ethical {{AI}} in healthcare: A
  five-facts approach. AI and Ethics \doi{10.1007/s43681-023-00353-x}

\bibitem[{Jackson(1995)}]{jackson1995world}
Jackson M (1995) The world and the machine. In: Proceedings of the 17th
  International Conference on Software Engineering. Association for Computing
  Machinery, New York, NY, USA, ICSE '95, p 283–292,
  \doi{10.1145/225014.225041}

\bibitem[{Jacobs and Wallach(2021)}]{jacobs2021measurement}
Jacobs AZ, Wallach H (2021) Measurement and fairness. In: Proceedings of the
  2021 ACM Conference on Fairness, Accountability, and Transparency.
  Association for Computing Machinery, New York, NY, USA, FAccT '21, p
  375–385, \doi{10.1145/3442188.3445901}

\bibitem[{Jakubovitz and Giryes(2018)}]{jakubovitz2018improving}
Jakubovitz D, Giryes R (2018) Improving {{DNN Robustness}} to {{Adversarial
  Attacks Using Jacobian Regularization}}. In: Ferrari V, Hebert M,
  Sminchisescu C, et~al (eds) Computer {{Vision}} \textendash{} {{ECCV}} 2018.
  {Springer International Publishing}, {Cham}, Lecture {{Notes}} in {{Computer
  Science}}, pp 525--541, \doi{10.1007/978-3-030-01258-8_32}

\bibitem[{Ji et~al(2023)Ji, Lee, Frieske, Yu, Su, Xu, Ishii, Bang, Madotto, and
  Fung}]{ji2023survey}
Ji Z, Lee N, Frieske R, et~al (2023) {Survey of Hallucination in Natural
  Language Generation}. ACM Comput Surv 55(12). \doi{10.1145/3571730}

\bibitem[{Jobin et~al(2019)Jobin, Ienca, and Vayena}]{jobin2019global}
Jobin A, Ienca M, Vayena E (2019) {The global landscape of AI ethics
  guidelines}. Nature Machine Intelligence 1(9):389--399

\bibitem[{Jung et~al(2022)Jung, Park, Chun, and Moon}]{jung2022Reweighting}
Jung S, Park T, Chun S, et~al (2022) Re-weighting {{Based Group Fairness
  Regularization}} via {{Classwise Robust Optimization}}. In: The {{Eleventh
  International Conference}} on {{Learning Representations}}

\bibitem[{Karpukhin et~al(2019)Karpukhin, Levy, Eisenstein, and
  Ghazvininejad}]{karpukhin2019Training}
Karpukhin V, Levy O, Eisenstein J, et~al (2019) Training on {{Synthetic Noise
  Improves Robustness}} to {{Natural Noise}} in {{Machine Translation}}. In:
  Proceedings of the 5th {{Workshop}} on {{Noisy User-generated Text}}
  ({{W-NUT}} 2019). {Association for Computational Linguistics}, {Hong Kong,
  China}, pp 42--47, \doi{10.18653/v1/D19-5506}

\bibitem[{Kasneci et~al(2023)Kasneci, Sessler, Küchemann, Bannert, Dementieva,
  Fischer, Gasser, Groh, Günnemann, Hüllermeier, Krusche, Kutyniok, Michaeli,
  Nerdel, Pfeffer, Poquet, Sailer, Schmidt, Seidel, Stadler, Weller, Kuhn, and
  Kasneci}]{Kasneci2023chatGPT}
Kasneci E, Sessler K, Küchemann S, et~al (2023) {ChatGPT for good? On
  opportunities and challenges of large language models for education}.
  Learning and Individual Differences 103:102274.
  \doi{https://doi.org/10.1016/j.lindif.2023.102274},
  \urlprefix\url{https://www.sciencedirect.com/science/article/pii/S1041608023000195}

\bibitem[{Kiden et~al(2024)Kiden, Stahl, Townsend, Maple, Vincent, Sampson,
  Gilbert, Smith, Deshmukh, Ross, Williams, {del Rincon}, Lisinska, O'Shea,
  Da~Costa~Abreu, Bencomo, Deb, Winter, Li, Torr, Lau, Iniesta, Ramchurn,
  Stein, and Yazdanpanah}]{kiden2024Responsible}
Kiden S, Stahl B, Townsend B, et~al (2024) Responsible {{AI}} governance: {{A}}
  response to {{UN}} interim report on governing {{AI}} for humanity.
  https://eprints.soton.ac.uk/488908/, \doi{10.5258/SOTON/PP0057}

\bibitem[{Kirichenko et~al(2022)Kirichenko, Izmailov, and
  Wilson}]{kirichenko2022Last}
Kirichenko P, Izmailov P, Wilson AG (2022) Last {{Layer Re-Training}} is
  {{Sufficient}} for {{Robustness}} to {{Spurious Correlations}}. In: The
  {{Eleventh International Conference}} on {{Learning Representations}}

\bibitem[{Kirk et~al(2022)Kirk, Vidgen, Rottger, Thrush, and
  Hale}]{kirk2022Hatemojia}
Kirk H, Vidgen B, Rottger P, et~al (2022) Hatemoji: {{A Test Suite}} and
  {{Adversarially-Generated Dataset}} for {{Benchmarking}} and {{Detecting
  Emoji-Based Hate}}. In: Proceedings of the 2022 {{Conference}} of the {{North
  American Chapter}} of the {{Association}} for {{Computational Linguistics}}:
  {{Human Language Technologies}}. {Association for Computational Linguistics},
  {Seattle, United States}, pp 1352--1368, \doi{10.18653/v1/2022.naacl-main.97}

\bibitem[{Kleinberg et~al(2017)Kleinberg, Mullainathan, and
  Raghavan}]{kleinberg2017inherent}
Kleinberg J, Mullainathan S, Raghavan M (2017) {Inherent Trade-Offs in the Fair
  Determination of Risk Scores}. In: Papadimitriou CH (ed) 8th Innovations in
  Theoretical Computer Science Conference ({ITCS} 2017), Leibniz International
  Proceedings in Informatics ({LIPIcs}), vol~67. Schloss
  Dagstuhl–Leibniz-Zentrum fuer Informatik, pp 43:1--43:23,
  \doi{10.4230/LIPIcs.ITCS.2017.43},
  \urlprefix\url{http://drops.dagstuhl.de/opus/volltexte/2017/8156}

\bibitem[{Komiyama et~al(2018)Komiyama, Takeda, Honda, and
  Shimao}]{komiyamaNonconvexOptimizationRegression2018}
Komiyama J, Takeda A, Honda J, et~al (2018) Nonconvex {{Optimization}} for
  {{Regression}} with {{Fairness Constraints}}. In: Proceedings of the 35th
  {{International Conference}} on {{Machine Learning}}. {PMLR}, pp 2737--2746

\bibitem[{Kononenko(2001)}]{kononenko2001machine}
Kononenko I (2001) Machine learning for medical diagnosis: history, state of
  the art and perspective. Artificial Intelligence in Medicine 23(1):89--109.
  \doi{https://doi.org/10.1016/S0933-3657(01)00077-X},
  \urlprefix\url{https://www.sciencedirect.com/science/article/pii/S093336570100077X}

\bibitem[{Krizhevsky et~al(2012)Krizhevsky, Sutskever, and
  Hinton}]{krizhevskz2012imagenet}
Krizhevsky A, Sutskever I, Hinton GE (2012) Imagenet classification with deep
  convolutional neural networks. In: Pereira F, Burges C, Bottou L, et~al (eds)
  Advances in Neural Information Processing Systems, vol~25. Curran Associates,
  Inc.,
  \urlprefix\url{https://proceedings.neurips.cc/paper_files/paper/2012/file/c399862d3b9d6b76c8436e924a68c45b-Paper.pdf}

\bibitem[{Lake and Baroni(2018)}]{lake2018generalization}
Lake B, Baroni M (2018) Generalization without systematicity: On the
  compositional skills of sequence-to-sequence recurrent networks. In: Dy J,
  Krause A (eds) Proceedings of the 35th International Conference on Machine
  Learning, Proceedings of Machine Learning Research, vol~80. PMLR, pp
  2873--2882, \urlprefix\url{https://proceedings.mlr.press/v80/lake18a.html}

\bibitem[{Lample et~al(2016)Lample, Ballesteros, Subramanian, Kawakami, and
  Dyer}]{lample2016neural}
Lample G, Ballesteros M, Subramanian S, et~al (2016) {Neural Architectures for
  Named Entity Recognition}. In: Knight K, Nenkova A, Rambow O (eds)
  Proceedings of the 2016 Conference of the North {A}merican Chapter of the
  Association for Computational Linguistics: Human Language Technologies.
  Association for Computational Linguistics, San Diego, California, pp
  260--270, \doi{10.18653/v1/N16-1030},
  \urlprefix\url{https://aclanthology.org/N16-1030}

\bibitem[{LeCun et~al(1989)LeCun, Boser, Denker, Henderson, Howard, Hubbard,
  and Jackel}]{lecun1989backpropagation}
LeCun Y, Boser B, Denker JS, et~al (1989) {Backpropagation Applied to
  Handwritten Zip Code Recognition}. Neural Computation 1(4):541--551.
  \doi{10.1162/neco.1989.1.4.541}

\bibitem[{Lee et~al(2022)Lee, Kim, Olfat, Hasegawa-Johnson, and
  Yoo}]{lee2022Fast}
Lee J, Kim G, Olfat M, et~al (2022) {Fast and Efficient MMD-Based Fair PCA via
  Optimization over Stiefel Manifold}. In: Proceedings of the AAAI Conference
  on Artificial Intelligence, pp 7363--7371, \doi{10.1609/aaai.v36i7.20699},
  \urlprefix\url{https://ojs.aaai.org/index.php/AAAI/article/view/20699}

\bibitem[{Levine et~al(2016)Levine, Finn, Darrell, and Abbeel}]{levine2016end}
Levine S, Finn C, Darrell T, et~al (2016) End-to-end training of deep
  visuomotor policies. Journal of Machine Learning Research 17(39):1--40.
  \urlprefix\url{http://jmlr.org/papers/v17/15-522.html}

\bibitem[{Levy et~al(2023)Levy, Bogin, and Berant}]{levy2023Diverse}
Levy I, Bogin B, Berant J (2023) Diverse {{Demonstrations Improve In-context
  Compositional Generalization}}. In: Rogers A, Boyd{-}Graber JL, Okazaki N
  (eds) Proceedings of the 61st {{Annual Meeting}} of the {{Association}} for
  {{Computational Linguistics}} ({{Volume}} 1: {{Long Papers}}). {Association
  for Computational Linguistics}, {Toronto, Canada}, pp 1401--1422,
  \doi{10.18653/v1/2023.acl-long.78}

\bibitem[{Ley(2002)}]{ley2002dblp}
Ley M (2002) {The DBLP computer science bibliography: Evolution, research
  issues, perspectives}. In: International symposium on string processing and
  information retrieval, Springer, pp 1--10

\bibitem[{Li et~al(2023{\natexlab{a}})Li, Guo, Zuo, and Chen}]{li2023squeeze}
Li Q, Guo Y, Zuo W, et~al (2023{\natexlab{a}}) {Squeeze Training for
  Adversarial Robustness}. In: The Eleventh International Conference on
  Learning Representations,
  \urlprefix\url{https://openreview.net/forum?id=Z_tmYu060Kr}

\bibitem[{Li et~al(2023{\natexlab{b}})Li, Wu, and Su}]{li2023Accurate}
Li X, Wu P, Su J (2023{\natexlab{b}}) Accurate fairness: Improving individual
  fairness without trading accuracy. In: Proceedings of the {{Thirty-Seventh
  AAAI Conference}} on {{Artificial Intelligence}} and {{Thirty-Fifth
  Conference}} on {{Innovative Applications}} of {{Artificial Intelligence}}
  and {{Thirteenth Symposium}} on {{Educational Advances}} in {{Artificial
  Intelligence}}, {{AAAI}}'23/{{IAAI}}'23/{{EAAI}}'23, vol~37. {AAAI Press}, pp
  14312--14320, \doi{10.1609/aaai.v37i12.26674}

\bibitem[{Liang et~al(2022)Liang, Sun, Zheng, and Huang}]{liang2022Efficient}
Liang Y, Sun Y, Zheng R, et~al (2022) Efficient {{Adversarial Training}}
  without {{Attacking}}: {{Worst-Case-Aware Robust Reinforcement Learning}}.
  In: Advances in {{Neural Information Processing Systems}}

\bibitem[{Lin et~al(2022)Lin, Hilton, and Evans}]{lin2022truthfulqa}
Lin S, Hilton J, Evans O (2022) {T}ruthful{QA}: Measuring how models mimic
  human falsehoods. In: Muresan S, Nakov P, Villavicencio A (eds) Proceedings
  of the 60th Annual Meeting of the Association for Computational Linguistics
  (Volume 1: Long Papers). Association for Computational Linguistics, Dublin,
  Ireland, pp 3214--3252, \doi{10.18653/v1/2022.acl-long.229},
  \urlprefix\url{https://aclanthology.org/2022.acl-long.229}

\bibitem[{Liu et~al(2021)Liu, Haghgoo, Chen, Raghunathan, Koh, Sagawa, Liang,
  and Finn}]{liu2021Just}
Liu EZ, Haghgoo B, Chen AS, et~al (2021) Just {{Train Twice}}: {{Improving
  Group Robustness}} without {{Training Group Information}}. In: Proceedings of
  the 38th {{International Conference}} on {{Machine Learning}}. {PMLR}, pp
  6781--6792

\bibitem[{Liu et~al(2019{\natexlab{a}})Liu, Schwartz, and
  Smith}]{liu2019Inoculation}
Liu NF, Schwartz R, Smith NA (2019{\natexlab{a}}) Inoculation by
  {{Fine-Tuning}}: {{A Method}} for {{Analyzing Challenge Datasets}}. In:
  Proceedings of the 2019 {{Conference}} of the {{North American Chapter}} of
  the {{Association}} for {{Computational Linguistics}}: {{Human Language
  Technologies}}, {{Volume}} 1 ({{Long}} and {{Short Papers}}). {Association
  for Computational Linguistics}, {Minneapolis, Minnesota}, pp 2171--2179,
  \doi{10.18653/v1/N19-1225}

\bibitem[{Liu et~al(2019{\natexlab{b}})Liu, Ott, Goyal, Du, Joshi, Chen, Levy,
  Lewis, Zettlemoyer, and Stoyanov}]{liu2019roberta}
Liu Y, Ott M, Goyal N, et~al (2019{\natexlab{b}}) {RoBERTa: {A} Robustly
  Optimized {BERT} Pretraining Approach}. CoRR abs/1907.11692.
  \urlprefix\url{http://arxiv.org/abs/1907.11692},
  {\href{https://arxiv.org/abs/1907.11692}{{1907.11692}}}

\bibitem[{L\"utjens et~al(2020)L\"utjens, Everett, and
  How}]{lutjens2020certified}
L\"utjens B, Everett M, How JP (2020) Certified adversarial robustness for deep
  reinforcement learning. In: Kaelbling LP, Kragic D, Sugiura K (eds)
  Proceedings of the Conference on Robot Learning, Proceedings of Machine
  Learning Research, vol 100. PMLR, pp 1328--1337,
  \urlprefix\url{https://proceedings.mlr.press/v100/lutjens20a.html}

\bibitem[{{Luz de Araujo} and {Roth}(2023)}]{luzdearaujo2023crossFunctional}
{Luz de Araujo} PH, {Roth} B (2023) {Cross-functional Analysis of
  Generalization in Behavioral Learning}. Transactions of the Association for
  Computational Linguistics 11:1066--1081. \doi{10.1162/tacl_a_00590},
  {\href{https://arxiv.org/abs/https://direct.mit.edu/tacl/article-pdf/doi/10.1162/tacl\_a\_00590/2154470/tacl\_a\_00590.pdf}{{https://direct.mit.edu/tacl/article-pdf/doi/10.1162/tacl\_a\_00590/2154470/tacl\_a\_00590.pdf}}}

\bibitem[{Ma et~al(2022)Ma, Ren, Zhao, Testuggine, and Peng}]{ma2022Are}
Ma M, Ren J, Zhao L, et~al (2022) Are {{Multimodal Transformers Robust}} to
  {{Missing Modality}}? In: Proceedings of the {{IEEE}}/{{CVF Conference}} on
  {{Computer Vision}} and {{Pattern Recognition}}, pp 18177--18186

\bibitem[{Madaio et~al(2020)Madaio, Stark, Wortman~Vaughan, and
  Wallach}]{madaio2020coDesigning}
Madaio MA, Stark L, Wortman~Vaughan J, et~al (2020) {Co-Designing Checklists to
  Understand Organizational Challenges and Opportunities around Fairness in
  AI}. In: Proceedings of the 2020 CHI Conference on Human Factors in Computing
  Systems. Association for Computing Machinery, New York, NY, USA, CHI '20, p
  1–14, \doi{10.1145/3313831.3376445}

\bibitem[{Madras et~al(2018)Madras, Pitassi, and
  Zemel}]{madrasPredictResponsiblyImproving2018}
Madras D, Pitassi T, Zemel R (2018) Predict {{Responsibly}}: {{Improving
  Fairness}} and {{Accuracy}} by {{Learning}} to {{Defer}}. In: Advances in
  {{Neural Information Processing Systems}}, vol~31. {Curran Associates, Inc.}

\bibitem[{Malik(2020)}]{malik2020hierarchy}
Malik MM (2020) A hierarchy of limitations in machine learning. CoRR
  abs/2002.05193. \urlprefix\url{https://arxiv.org/abs/2002.05193}

\bibitem[{Mehrabi et~al(2021)Mehrabi, Morstatter, Saxena, Lerman, and
  Galstyan}]{mehrabi2021survey}
Mehrabi N, Morstatter F, Saxena N, et~al (2021) {A Survey on Bias and Fairness
  in Machine Learning}. ACM Comput Surv 54(6). \doi{10.1145/3457607}

\bibitem[{Min et~al(2020)Min, McCoy, Das, Pitler, and
  Linzen}]{min2020Syntactic}
Min J, McCoy RT, Das D, et~al (2020) Syntactic {{Data Augmentation Increases
  Robustness}} to {{Inference Heuristics}}. In: Proceedings of the 58th
  {{Annual Meeting}} of the {{Association}} for {{Computational Linguistics}}.
  {Association for Computational Linguistics}, {Online}, pp 2339--2352,
  \doi{10.18653/v1/2020.acl-main.212}

\bibitem[{Minaee et~al(2022)Minaee, Boykov, Porikli, Plaza, Kehtarnavaz, and
  Terzopoulos}]{minaee2022image}
Minaee S, Boykov Y, Porikli F, et~al (2022) {Image Segmentation Using Deep
  Learning: A Survey}. IEEE Transactions on Pattern Analysis and Machine
  Intelligence 44(7):3523--3542. \doi{10.1109/TPAMI.2021.3059968}

\bibitem[{Mishler et~al(2021)Mishler, Kennedy, and
  Chouldechova}]{mishler2021Fairness}
Mishler A, Kennedy EH, Chouldechova A (2021) Fairness in {{Risk Assessment
  Instruments}}: {{Post-Processing}} to {{Achieve Counterfactual Equalized
  Odds}}. In: Proceedings of the 2021 {{ACM Conference}} on {{Fairness}},
  {{Accountability}}, and {{Transparency}}. {Association for Computing
  Machinery}, {New York, NY, USA}, {{FAccT}} '21, pp 386--400,
  \doi{10.1145/3442188.3445902}

\bibitem[{Mnih et~al(2015)Mnih, Kavukcuoglu, Silver, Rusu, Veness, Bellemare,
  Graves, Riedmiller, Fidjeland, Ostrovski, Petersen, Beattie, Sadik,
  Antonoglou, King, Kumaran, Wierstra, Legg, and Hassabis}]{mnih2015human}
Mnih V, Kavukcuoglu K, Silver D, et~al (2015) Human-level control through deep
  reinforcement learning. Nature 518(7540):529--533. \doi{10.1038/nature14236},
  \urlprefix\url{https://doi.org/10.1038/nature14236}

\bibitem[{Naik et~al(2018)Naik, Ravichander, Sadeh, Rose, and
  Neubig}]{naik2018stress}
Naik A, Ravichander A, Sadeh N, et~al (2018) Stress {{Test Evaluation}} for
  {{Natural Language Inference}}. In: Proceedings of the 27th {{International
  Conference}} on {{Computational Linguistics}}. {Association for Computational
  Linguistics}, {Santa Fe, New Mexico, USA}, pp 2340--2353

\bibitem[{Nangia et~al(2020)Nangia, Vania, Bhalerao, and
  Bowman}]{nangia2020crows}
Nangia N, Vania C, Bhalerao R, et~al (2020) {{C}row{S}-Pairs: A Challenge
  Dataset for Measuring Social Biases in Masked Language Models}. In: Webber B,
  Cohn T, He Y, et~al (eds) Proceedings of the 2020 Conference on Empirical
  Methods in Natural Language Processing (EMNLP). Association for Computational
  Linguistics, Online, pp 1953--1967, \doi{10.18653/v1/2020.emnlp-main.154},
  \urlprefix\url{https://aclanthology.org/2020.emnlp-main.154}

\bibitem[{Narasimhan et~al(2019)Narasimhan, Cotter, and
  Gupta}]{narasimhan2019Optimizing}
Narasimhan H, Cotter A, Gupta M (2019) Optimizing {{Generalized Rate Metrics}}
  with {{Three Players}}. In: Advances in {{Neural Information Processing
  Systems}}, vol~32. {Curran Associates, Inc.}

\bibitem[{OECD(2019)}]{OECD2019ai}
OECD (2019) {OECD AI Principles overview}. OECD AI Policy Observatory
  \urlprefix\url{https://oecd.ai/en/ai-principles}

\bibitem[{Ouyang et~al(2022)Ouyang, Wu, Jiang, Almeida, Wainwright, Mishkin,
  Zhang, Agarwal, Slama, Ray, Schulman, Hilton, Kelton, Miller, Simens, Askell,
  Welinder, Christiano, Leike, and Lowe}]{ouyang2022training}
Ouyang L, Wu J, Jiang X, et~al (2022) Training language models to follow
  instructions with human feedback. In: Koyejo S, Mohamed S, Agarwal A, et~al
  (eds) Advances in Neural Information Processing Systems, vol~35. Curran
  Associates, Inc., pp 27730--27744,
  \urlprefix\url{https://proceedings.neurips.cc/paper_files/paper/2022/file/b1efde53be364a73914f58805a001731-Paper-Conference.pdf}

\bibitem[{Pessach and Shmueli(2023)}]{pessach2023review}
Pessach D, Shmueli E (2023) A {{Review}} on {{Fairness}} in {{Machine
  Learning}}. ACM Computing Surveys 55(3):1--44. \doi{10.1145/3494672}

\bibitem[{Petersen et~al(2023)Petersen, Ganz, Holm, and
  Feragen}]{petersen2023assessing}
Petersen E, Ganz M, Holm S, et~al (2023) On (assessing) the fairness of risk
  score models. In: Proceedings of the 2023 {{ACM Conference}} on {{Fairness}},
  {{Accountability}}, and {{Transparency}}. {Association for Computing
  Machinery}, {New York, NY, USA}, {{FAccT}} '23, pp 817--829,
  \doi{10.1145/3593013.3594045}

\bibitem[{Pfohl et~al(2022{\natexlab{a}})Pfohl, Xu, Foryciarz, Ignatiadis,
  Genkins, and Shah}]{pfohl2022Net}
Pfohl S, Xu Y, Foryciarz A, et~al (2022{\natexlab{a}}) {Net Benefit,
  Calibration, Threshold Selection, and Training Objectives for Algorithmic
  Fairness in Healthcare}. In: Proceedings of the 2022 {{ACM Conference}} on
  {{Fairness}}, {{Accountability}}, and {{Transparency}}. {Association for
  Computing Machinery}, {New York, NY, USA}, {{FAccT}} '22, pp 1039--1052,
  \doi{10.1145/3531146.3533166}

\bibitem[{Pfohl et~al(2022{\natexlab{b}})Pfohl, Zhang, Xu, Foryciarz, Ghassemi,
  and Shah}]{pfohl2022comparison}
Pfohl SR, Zhang H, Xu Y, et~al (2022{\natexlab{b}}) A comparison of approaches
  to improve worst-case predictive model performance over patient
  subpopulations. Scientific reports 12(1):3254

\bibitem[{Qiu et~al(2022)Qiu, Shaw, Pasupat, Nowak, Linzen, Sha, and
  Toutanova}]{qiu2022Improving}
Qiu L, Shaw P, Pasupat P, et~al (2022) Improving {{Compositional
  Generalization}} with {{Latent Structure}} and {{Data Augmentation}}. In:
  Proceedings of the 2022 {{Conference}} of the {{North American Chapter}} of
  the {{Association}} for {{Computational Linguistics}}: {{Human Language
  Technologies}}. {Association for Computational Linguistics}, {Seattle, United
  States}, pp 4341--4362, \doi{10.18653/v1/2022.naacl-main.323}

\bibitem[{Radford et~al(2021)Radford, Kim, Hallacy, Ramesh, Goh, Agarwal,
  Sastry, Askell, Mishkin, Clark, Krueger, and Sutskever}]{radford2021learning}
Radford A, Kim JW, Hallacy C, et~al (2021) Learning transferable visual models
  from natural language supervision. In: Meila M, Zhang T (eds) Proceedings of
  the 38th International Conference on Machine Learning, Proceedings of Machine
  Learning Research, vol 139. PMLR, pp 8748--8763,
  \urlprefix\url{https://proceedings.mlr.press/v139/radford21a.html}

\bibitem[{Raffel et~al(2020)Raffel, Shazeer, Roberts, Lee, Narang, Matena,
  Zhou, Li, and Liu}]{raffel2020exploring}
Raffel C, Shazeer N, Roberts A, et~al (2020) Exploring the limits of transfer
  learning with a unified text-to-text transformer. Journal of Machine Learning
  Research 21(140):1--67.
  \urlprefix\url{http://jmlr.org/papers/v21/20-074.html}

\bibitem[{Rahmattalabi et~al(2021)Rahmattalabi, Jabbari, Lakkaraju, Vayanos,
  Izenberg, Brown, Rice, and Tambe}]{rahmattalabi2020Fair}
Rahmattalabi A, Jabbari S, Lakkaraju H, et~al (2021) Fair influence
  maximization: a welfare optimization approach. In: Proceedings of the AAAI
  Conference on Artificial Intelligence, pp 11630--11638,
  \doi{10.1609/aaai.v35i13.17383},
  \urlprefix\url{https://ojs.aaai.org/index.php/AAAI/article/view/17383}

\bibitem[{Ramesh et~al(2021)Ramesh, Pavlov, Goh, Gray, Voss, Radford, Chen, and
  Sutskever}]{ramesh2021zero}
Ramesh A, Pavlov M, Goh G, et~al (2021) {Zero-Shot Text-to-Image Generation}.
  In: Meila M, Zhang T (eds) Proceedings of the 38th International Conference
  on Machine Learning, Proceedings of Machine Learning Research, vol 139. PMLR,
  pp 8821--8831,
  \urlprefix\url{https://proceedings.mlr.press/v139/ramesh21a.html}

\bibitem[{Rebuffi et~al(2021)Rebuffi, Gowal, Calian, Stimberg, Wiles, and
  Mann}]{rebuffi2021Data}
Rebuffi SA, Gowal S, Calian DA, et~al (2021) Data {{Augmentation Can Improve
  Robustness}}. In: Advances in {{Neural Information Processing Systems}},
  \eprint{2111.05328}

\bibitem[{Ribeiro et~al(2016)Ribeiro, Singh, and Guestrin}]{ribeiro2016should}
Ribeiro MT, Singh S, Guestrin C (2016) Why should {I} trust you?: Explaining
  the predictions of any classifier. In: Proceedings of the 22nd ACM SIGKDD
  international conference on knowledge discovery and data mining, ACM, pp
  1135--1144

\bibitem[{Ribeiro et~al(2020)Ribeiro, Wu, Guestrin, and
  Singh}]{ribeiro2020beyond}
Ribeiro MT, Wu T, Guestrin C, et~al (2020) Beyond {{Accuracy}}: {{Behavioral
  Testing}} of {{NLP Models}} with {{CheckList}}. In: Proceedings of the 58th
  {{Annual Meeting}} of the {{Association}} for {{Computational Linguistics}}.
  {Association for Computational Linguistics}, {Online}, pp 4902--4912,
  \doi{10.18653/v1/2020.acl-main.442}

\bibitem[{Roelofs et~al(2019)Roelofs, Shankar, Recht, Fridovich-Keil, Hardt,
  Miller, and Schmidt}]{roelofs2019meta}
Roelofs R, Shankar V, Recht B, et~al (2019) A meta-analysis of overfitting in
  machine learning. Advances in Neural Information Processing Systems 32

\bibitem[{Roh et~al(2021)Roh, Lee, Whang, and Suh}]{roh2021Sample}
Roh Y, Lee K, Whang SE, et~al (2021) Sample {{Selection}} for {{Fair}} and
  {{Robust Training}}. In: Advances in {{Neural Information Processing
  Systems}}

\bibitem[{Ross and {Doshi-Velez}(2018)}]{ross2018improving}
Ross A, {Doshi-Velez} F (2018) Improving the {{Adversarial Robustness}} and
  {{Interpretability}} of {{Deep Neural Networks}} by {{Regularizing Their
  Input Gradients}}. In: Proceedings of the AAAI Conference on Artificial
  Intelligence, \doi{10.1609/aaai.v32i1.11504}

\bibitem[{R{\"o}ttger et~al(2021)R{\"o}ttger, Vidgen, Nguyen, Waseem, Margetts,
  and Pierrehumbert}]{rottger2021hatecheck}
R{\"o}ttger P, Vidgen B, Nguyen D, et~al (2021) {{HateCheck}}: {{Functional
  Tests}} for {{Hate Speech Detection Models}}. In: Proceedings of the 59th
  {{Annual Meeting}} of the {{Association}} for {{Computational Linguistics}}
  and the 11th {{International Joint Conference}} on {{Natural Language
  Processing}} ({{Volume}} 1: {{Long Papers}}). {Association for Computational
  Linguistics}, {Online}, pp 41--58, \doi{10.18653/v1/2021.acl-long.4}

\bibitem[{Ruis et~al(2020)Ruis, Andreas, Baroni, Bouchacourt, and
  Lake}]{ruis2020Benchmark}
Ruis L, Andreas J, Baroni M, et~al (2020) A {{Benchmark}} for {{Systematic
  Generalization}} in {{Grounded Language Understanding}}. In: Advances in
  {{Neural Information Processing Systems}}, vol~33. {Curran Associates, Inc.},
  pp 19861--19872

\bibitem[{Russakovsky et~al(2015)Russakovsky, Deng, Su, Krause, Satheesh, Ma,
  Huang, Karpathy, Khosla, Bernstein, Berg, and
  Fei-Fei}]{russakovsky2015imagenet}
Russakovsky O, Deng J, Su H, et~al (2015) {ImageNet Large Scale Visual
  Recognition Challenge}. International Journal of Computer Vision
  115(3):211--252. \doi{10.1007/s11263-015-0816-y}

\bibitem[{Schick et~al(2021)Schick, Udupa, and
  Schütze}]{schick2021selfdiagnosis}
Schick T, Udupa S, Schütze H (2021) {Self-Diagnosis and Self-Debiasing: A
  Proposal for Reducing Corpus-Based Bias in NLP}. Transactions of the
  Association for Computational Linguistics 9:1408--1424.
  \doi{10.1162/tacl_a_00434},
  {\href{https://arxiv.org/abs/https://direct.mit.edu/tacl/article-pdf/doi/10.1162/tacl\_a\_00434/1979270/tacl\_a\_00434.pdf}{{https://direct.mit.edu/tacl/article-pdf/doi/10.1162/tacl\_a\_00434/1979270/tacl\_a\_00434.pdf}}}

\bibitem[{Schneider et~al(2020)Schneider, Rusak, Eck, Bringmann, Brendel, and
  Bethge}]{schneider2020Improving}
Schneider S, Rusak E, Eck L, et~al (2020) {Improving Robustness against Common
  Corruptions by Covariate Shift Adaptation}. In: Advances in {{Neural
  Information Processing Systems}}, vol~33. {Curran Associates, Inc.}, pp
  11539--11551

\bibitem[{Sehwag et~al(2022)Sehwag, Mahloujifar, Handina, Dai, Xiang, Chiang,
  and Mittal}]{sehwag2022Robust}
Sehwag V, Mahloujifar S, Handina T, et~al (2022) Robust {{Learning Meets
  Generative Models}}: {{Can Proxy Distributions Improve Adversarial
  Robustness}}? In: The {{Tenth International Conference}} on {{Learning
  Representations}}, {{ICLR}} 2022, {{Virtual Event}}, {{April}} 25-29, 2022.
  {OpenReview.net}

\bibitem[{Shalev-Shwartz and Ben-David(2014)}]{shalev2014understanding}
Shalev-Shwartz S, Ben-David S (2014) Understanding Machine Learning: From
  Theory to Algorithms. Cambridge University Press

\bibitem[{Sinha et~al(2019)Sinha, Namkoong, and
  Duchi}]{sinhaCertifyingDistributionalRobustness2019}
Sinha A, Namkoong H, Duchi J (2019) Certifying {{Some Distributional
  Robustness}} with {{Principled Adversarial Training}}. In: International
  {{Conference}} on {{Learning Representations}}

\bibitem[{Skalse et~al(2022)Skalse, Howe, Krasheninnikov, and
  Krueger}]{skalse2022defining}
Skalse JMV, Howe NHR, Krasheninnikov D, et~al (2022) Defining and
  characterizing reward gaming. In: Oh AH, Agarwal A, Belgrave D, et~al (eds)
  Advances in Neural Information Processing Systems,
  \urlprefix\url{https://openreview.net/forum?id=yb3HOXO3lX2}

\bibitem[{Socher et~al(2013)Socher, Perelygin, Wu, Chuang, Manning, Ng, and
  Potts}]{socher2013recursive}
Socher R, Perelygin A, Wu J, et~al (2013) {Recursive Deep Models for Semantic
  Compositionality Over a Sentiment Treebank}. In: Yarowsky D, Baldwin T,
  Korhonen A, et~al (eds) Proceedings of the 2013 Conference on Empirical
  Methods in Natural Language Processing. Association for Computational
  Linguistics, Seattle, Washington, USA, pp 1631--1642,
  \urlprefix\url{https://aclanthology.org/D13-1170}

\bibitem[{Sreenu and Saleem~Durai(2019)}]{sreenu2019intelligent}
Sreenu G, Saleem~Durai MA (2019) Intelligent video surveillance: a review
  through deep learning techniques for crowd analysis. Journal of Big Data
  6(1):48. \doi{10.1186/s40537-019-0212-5}

\bibitem[{Sun et~al(2023)Sun, Dou, Yang, Zhang, Wang, Yu, He, and
  Li}]{sun2023adversarial}
Sun L, Dou Y, Yang C, et~al (2023) {Adversarial Attack and Defense on Graph
  Data: A Survey}. IEEE Transactions on Knowledge and Data Engineering
  35(8):7693--7711. \doi{10.1109/TKDE.2022.3201243}

\bibitem[{Sun et~al(2020)Sun, Wang, Liu, Miller, Efros, and
  Hardt}]{sun2020TestTime}
Sun Y, Wang X, Liu Z, et~al (2020) Test-{{Time Training}} with
  {{Self-Supervision}} for {{Generalization}} under {{Distribution Shifts}}.
  In: Proceedings of the 37th {{International Conference}} on {{Machine
  Learning}}. {PMLR}, pp 9229--9248

\bibitem[{Sutton and Barto(2018)}]{sutton2018reinforcement}
Sutton RS, Barto AG (2018) Reinforcement Learning: An Introduction. A Bradford
  Book, Cambridge, MA, USA

\bibitem[{Szegedy et~al(2016)Szegedy, Vanhoucke, Ioffe, Shlens, and
  Wojna}]{szegedy2016rethinking}
Szegedy C, Vanhoucke V, Ioffe S, et~al (2016) {Rethinking the Inception
  Architecture for Computer Vision}. In: 2016 IEEE Conference on Computer
  Vision and Pattern Recognition (CVPR), pp 2818--2826,
  \doi{10.1109/CVPR.2016.308}

\bibitem[{Taskesen et~al(2021)Taskesen, Blanchet, Kuhn, and
  Nguyen}]{taskesen2021Statistical}
Taskesen B, Blanchet J, Kuhn D, et~al (2021) A {{Statistical Test}} for
  {{Probabilistic Fairness}}. In: Proceedings of the 2021 {{ACM Conference}} on
  {{Fairness}}, {{Accountability}}, and {{Transparency}}. {Association for
  Computing Machinery}, {New York, NY, USA}, {{FAccT}} '21, pp 648--665,
  \doi{10.1145/3442188.3445927}

\bibitem[{{The European Comission}(2003)}]{EC2003guidelines}
{The European Comission} (2003) {General Guidelines for the Cooperation between
  CEN, Cenelec and ETSI and the European Commission and the European Free Trade
  Association}.
  \urlprefix\url{https://eur-lex.europa.eu/legal-content/EN/ALL/?uri=CELEX:52003XC0416(03)}

\bibitem[{{The European Comission}(2008)}]{EC2008new}
{The European Comission} (2008) New legislative framework.
  \urlprefix\url{https://single-market-economy.ec.europa.eu/single-market/goods/new-legislative-framework_en}

\bibitem[{{The European Comission}(2018)}]{EC2018factsheet}
{The European Comission} (2018) {Factsheet: Artificial Intelligence for
  Europe}.
  \urlprefix\url{https://digital-strategy.ec.europa.eu/en/library/ethics-guidelines-trustworthy-ai}

\bibitem[{Thirunavukarasu et~al(2023)Thirunavukarasu, Ting, Elangovan,
  Gutierrez, Tan, and Ting}]{thirunavukarasu2023large}
Thirunavukarasu AJ, Ting DSJ, Elangovan K, et~al (2023) Large language models
  in medicine. Nature Medicine 29(8):1930--1940.
  \doi{10.1038/s41591-023-02448-8}

\bibitem[{Tjeng et~al(2019)Tjeng, Xiao, and Tedrake}]{tjeng2019Evaluating}
Tjeng V, Xiao KY, Tedrake R (2019) Evaluating {{Robustness}} of {{Neural
  Networks}} with {{Mixed Integer Programming}}. In: 7th {{International
  Conference}} on {{Learning Representations}}, {{ICLR}} 2019, {{New Orleans}},
  {{LA}}, {{USA}}, {{May}} 6-9, 2019. {OpenReview.net}

\bibitem[{Tu et~al(2020)Tu, Lalwani, Gella, and He}]{tu2020empirical}
Tu L, Lalwani G, Gella S, et~al (2020) {An Empirical Study on Robustness to
  Spurious Correlations using Pre-trained Language Models}. Transactions of the
  Association for Computational Linguistics 8:621--633.
  \doi{10.1162/tacl_a_00335},
  {\href{https://arxiv.org/abs/https://direct.mit.edu/tacl/article-pdf/doi/10.1162/tacl\_a\_00335/1923506/tacl\_a\_00335.pdf}{{https://direct.mit.edu/tacl/article-pdf/doi/10.1162/tacl\_a\_00335/1923506/tacl\_a\_00335.pdf}}}

\bibitem[{Vaswani et~al(2017)Vaswani, Shazeer, Parmar, Uszkoreit, Jones, Gomez,
  Kaiser, and Polosukhin}]{vaswani2017attention}
Vaswani A, Shazeer N, Parmar N, et~al (2017) Attention is all you need. In:
  Advances in neural information processing systems, pp 5998--6008

\bibitem[{Veale and Borgesius(2021)}]{veale2021demystifying}
Veale M, Borgesius FZ (2021) {Demystifying the Draft EU Artificial Intelligence
  Act—Analysing the good, the bad, and the unclear elements of the proposed
  approach}. Computer Law Review International 22(4):97--112

\bibitem[{Wachter et~al(2017)Wachter, Mittelstadt, and
  Russell}]{wachter2017counterfactual}
Wachter S, Mittelstadt B, Russell C (2017) {Counterfactual explanations without
  opening the black box: Automated decisions and the GDPR}. Harv JL \& Tech
  31:841

\bibitem[{Wang et~al(2020{\natexlab{a}})Wang, Wang, Cheng, Gan, Jia, Li, and
  Liu}]{wang2020InfoBERT}
Wang B, Wang S, Cheng Y, et~al (2020{\natexlab{a}}) {{InfoBERT}}: {{Improving
  Robustness}} of {{Language Models}} from {{An Information Theoretic
  Perspective}}. In: International {{Conference}} on {{Learning
  Representations}}

\bibitem[{Wang et~al(2020{\natexlab{b}})Wang, Guo, Narasimhan, Cotter, Gupta,
  and Jordan}]{wang2020Robust}
Wang S, Guo W, Narasimhan H, et~al (2020{\natexlab{b}}) Robust {{Optimization}}
  for {{Fairness}} with {{Noisy Protected Groups}}. In: Advances in {{Neural
  Information Processing Systems}}, vol~33. {Curran Associates, Inc.}, pp
  5190--5203

\bibitem[{Wang et~al(2022{\natexlab{a}})Wang, Sridhar, Yang, and
  Wang}]{wang2022Identifying}
Wang T, Sridhar R, Yang D, et~al (2022{\natexlab{a}}) Identifying and
  {{Mitigating Spurious Correlations}} for {{Improving Robustness}} in {{NLP
  Models}}. In: Findings of the {{Association}} for {{Computational
  Linguistics}}: {{NAACL}} 2022. {Association for Computational Linguistics},
  {Seattle, United States}, pp 1719--1729,
  \doi{10.18653/v1/2022.findings-naacl.130}

\bibitem[{Wang et~al(2022{\natexlab{b}})Wang, Wang, and Yang}]{wang2022measure}
Wang X, Wang H, Yang D (2022{\natexlab{b}}) Measure and improve robustness in
  {NLP} models: A survey. In: Proceedings of the 2022 Conference of the North
  American Chapter of the Association for Computational Linguistics: Human
  Language Technologies. Association for Computational Linguistics, Seattle,
  United States, pp 4569--4586, \doi{10.18653/v1/2022.naacl-main.339},
  \urlprefix\url{https://aclanthology.org/2022.naacl-main.339}

\bibitem[{Wang and Bansal(2018)}]{wangRobustMachineComprehension2018}
Wang Y, Bansal M (2018) Robust {{Machine Comprehension Models}} via
  {{Adversarial Training}}. In: Proceedings of the 2018 {{Conference}} of the
  {{North American Chapter}} of the {{Association}} for {{Computational
  Linguistics}}: {{Human Language Technologies}}, {{Volume}} 2 ({{Short
  Papers}}). {Association for Computational Linguistics}, {New Orleans,
  Louisiana}, pp 575--581, \doi{10.18653/v1/N18-2091}

\bibitem[{Wang et~al(2019)Wang, Zou, Yi, Bailey, Ma, and
  Gu}]{wang2019Improving}
Wang Y, Zou D, Yi J, et~al (2019) Improving {{Adversarial Robustness Requires
  Revisiting Misclassified Examples}}. In: International {{Conference}} on
  {{Learning Representations}}

\bibitem[{Webb(2010)}]{Webb2010}
Webb GI (2010) Overfitting, Springer US, Boston, MA, pp 744--744.
  \doi{10.1007/978-0-387-30164-8_623},
  \urlprefix\url{https://doi.org/10.1007/978-0-387-30164-8_623}

\bibitem[{Wei et~al(2022{\natexlab{a}})Wei, Bosma, Zhao, Guu, Yu, Lester, Du,
  Dai, and Le}]{wei2022finetuned}
Wei J, Bosma M, Zhao V, et~al (2022{\natexlab{a}}) Finetuned language models
  are zero-shot learners. In: International Conference on Learning
  Representations, \urlprefix\url{https://openreview.net/forum?id=gEZrGCozdqR}

\bibitem[{Wei et~al(2022{\natexlab{b}})Wei, Tay, Bommasani, Raffel, Zoph,
  Borgeaud, Yogatama, Bosma, Zhou, Metzler, Chi, Hashimoto, Vinyals, Liang,
  Dean, and Fedus}]{wei2022emergent}
Wei J, Tay Y, Bommasani R, et~al (2022{\natexlab{b}}) Emergent abilities of
  large language models. Transactions on Machine Learning Research
  \urlprefix\url{https://openreview.net/forum?id=yzkSU5zdwD}

\bibitem[{Weng et~al(2018)Weng, Zhang, Chen, Yi, Su, Gao, Hsieh, and
  Daniel}]{weng2018clever}
Weng TW, Zhang H, Chen PY, et~al (2018) Evaluating the robustness of neural
  networks: An extreme value theory approach. In: International Conference on
  Learning Representations (ICLR)

\bibitem[{Wiener(1960)}]{wiener1960some}
Wiener N (1960) {Some Moral and Technical Consequences of Automation: As
  machines learn they may develop unforeseen strategies at rates that baffle
  their programmers.} Science 131(3410):1355--1358

\bibitem[{Wu et~al(2020)Wu, Jiang, Ba, and Grosse}]{wu2020INT}
Wu Y, Jiang A, Ba J, et~al (2020) {{INT}}: {{An Inequality Benchmark}} for
  {{Evaluating Generalization}} in {{Theorem Proving}}. In: International
  {{Conference}} on {{Learning Representations}}

\bibitem[{Wu et~al(2021)Wu, Pan, Chen, Long, Zhang, and
  Yu}]{wu2021comprehensive}
Wu Z, Pan S, Chen F, et~al (2021) A comprehensive survey on graph neural
  networks. IEEE Transactions on Neural Networks and Learning Systems
  32(1):4--24. \doi{10.1109/TNNLS.2020.2978386}

\bibitem[{Xie et~al(2019)Xie, Wu, van~der Maaten, Yuille, and
  He}]{xie2019Feature}
Xie C, Wu Y, van~der Maaten L, et~al (2019) Feature {{Denoising}} for
  {{Improving Adversarial Robustness}}. In: Proceedings of the {{IEEE}}/{{CVF
  Conference}} on {{Computer Vision}} and {{Pattern Recognition}}, pp 501--509

\bibitem[{Xiong et~al(2020)Xiong, Wang, Liu, Zhong, Wan, Li, Li, Luo, Chen,
  Jiang, and Zheng}]{xiong2020pushing}
Xiong Z, Wang D, Liu X, et~al (2020) {Pushing the Boundaries of Molecular
  Representation for Drug Discovery with the Graph Attention Mechanism}.
  Journal of Medicinal Chemistry 63(16):8749--8760.
  \doi{10.1021/acs.jmedchem.9b00959}, pMID: 31408336

\bibitem[{Yuan et~al(2022)Yuan, Coenen, Reif, and Ippolito}]{yuan2022wordcraft}
Yuan A, Coenen A, Reif E, et~al (2022) {Wordcraft: Story Writing With Large
  Language Models}. In: 27th International Conference on Intelligent User
  Interfaces. Association for Computing Machinery, New York, NY, USA, IUI '22,
  p 841–852, \doi{10.1145/3490099.3511105}

\bibitem[{Zan et~al(2023)Zan, Chen, Zhang, Lu, Wu, Guan, Yongji, and
  Lou}]{zan2023large}
Zan D, Chen B, Zhang F, et~al (2023) Large language models meet {NL}2{C}ode: A
  survey. In: Rogers A, Boyd-Graber J, Okazaki N (eds) Proceedings of the 61st
  Annual Meeting of the Association for Computational Linguistics (Volume 1:
  Long Papers). Association for Computational Linguistics, Toronto, Canada, pp
  7443--7464, \doi{10.18653/v1/2023.acl-long.411},
  \urlprefix\url{https://aclanthology.org/2023.acl-long.411}

\bibitem[{Zhang et~al(2022{\natexlab{a}})Zhang, Tian, Ju, Liu, Ye, Chawla, and
  Zhang}]{zhang2022Chasing}
Zhang C, Tian Y, Ju M, et~al (2022{\natexlab{a}}) Chasing {{All-Round Graph
  Representation Robustness}}: {{Model}}, {{Training}}, and {{Optimization}}.
  In: The {{Eleventh International Conference}} on {{Learning Representations}}

\bibitem[{Zhang et~al(2019)Zhang, Chen, Xiao, Gowal, Stanforth, Li, Boning, and
  Hsieh}]{zhang2019Stable}
Zhang H, Chen H, Xiao C, et~al (2019) Towards {{Stable}} and {{Efficient
  Training}} of {{Verifiably Robust Neural Networks}}. In: International
  {{Conference}} on {{Learning Representations}}

\bibitem[{Zhang et~al(2022{\natexlab{b}})Zhang, Levine, and
  Finn}]{zhang2022MEMO}
Zhang M, Levine S, Finn C (2022{\natexlab{b}}) {{MEMO}}: {{Test Time
  Robustness}} via {{Adaptation}} and {{Augmentation}}. Advances in Neural
  Information Processing Systems 35:38629--38642

\bibitem[{Zhang et~al(2022{\natexlab{c}})Zhang, Sohoni, Zhang, Finn, and
  Re}]{zhang2022CorrectNContrast}
Zhang M, Sohoni NS, Zhang HR, et~al (2022{\natexlab{c}})
  Correct-{{N-Contrast}}: A {{Contrastive Approach}} for {{Improving
  Robustness}} to {{Spurious Correlations}}. In: Proceedings of the 39th
  {{International Conference}} on {{Machine Learning}}. {PMLR}, pp
  26484--26516, \eprint{2203.01517}

\bibitem[{Zhang et~al(2020)Zhang, Sheng, Alhazmi, and
  Li}]{zhang2020adversarial}
Zhang WE, Sheng QZ, Alhazmi A, et~al (2020) Adversarial attacks on
  deep-learning models in natural language processing: A survey. ACM Trans
  Intell Syst Technol 11(3). \doi{10.1145/3374217}

\bibitem[{Zhuo et~al(2023)Zhuo, Li, Huang, Shiri, Wang, Haffari, and
  Li}]{zhuo2023Robustness}
Zhuo TY, Li Z, Huang Y, et~al (2023) On {{Robustness}} of {{Prompt-based
  Semantic Parsing}} with {{Large Pre-trained Language Model}}: {{An Empirical
  Study}} on {{Codex}}. In: Vlachos A, Augenstein I (eds) Proceedings of the
  17th {{Conference}} of the {{European Chapter}} of the {{Association}} for
  {{Computational Linguistics}}. {Association for Computational Linguistics},
  {Dubrovnik, Croatia}, pp 1090--1102, \doi{10.18653/v1/2023.eacl-main.77}

\end{thebibliography}

\end{document}